\documentclass{article}

\usepackage{arxiv}

\usepackage[utf8]{inputenc} 
\usepackage[T1]{fontenc}    
\usepackage{hyperref}       
\usepackage{url}            
\usepackage{booktabs}       
\usepackage{amsfonts}       
\usepackage{nicefrac}       
\usepackage{microtype}      
\usepackage{lipsum}

\usepackage{graphicx}
\usepackage[load-configurations=version-1]{siunitx} 
\usepackage{multirow}
\usepackage{hyperref}
\usepackage{adjustbox}
\usepackage{amsmath}
\usepackage{cleveref}

\usepackage{floatrow}
\hypersetup{
    colorlinks=true,
    linkcolor=black,
    citecolor=black,
    urlcolor=blue,
}

\newcommand{\subthreesection}[1]{\paragraph{#1}\mbox{}\\}
\title{Measuring Sample Efficiency and Generalization in Reinforcement Learning Benchmarks: NeurIPS 2020 Procgen Benchmark}

\author{
  Sharada Mohanty \\
  AIcrowd Research, AIcrowd\\
  \texttt{mohanty@aicrowd.com} \\
 \And
 Jyotish Poonganam \\
 AIcrowd Research, AIcrowd\\
 \texttt{jyotish@aicrowd.com} \\

 \And
 Adrien Gaidon \\
 Toyota Research Institute\\
 \texttt{adrien.gaidon@tri.global} \\

 \And
 Andrey Kolobov \\
 Microsoft Research\\
 \texttt{akolobov@microsoft.com} \\

 \And
 Blake Wulfe \\
 Toyota Research Institute\\
 \texttt{blake.wulfe@tri.global} \\

 \And
 Dipam Chakraborty \\
 AIcrowd Research, AIcrowd\\
 \texttt{dipam@aicrowd.com} \\

 \And
 Gražvydas Šemetulskis \\
 Three Thirds\\
 \texttt{grazvydas@threethirds.ai} \\

 \And
 João Schapke \\
 Institute of Informatics, Universidade Federal do Rio Grande do Sul, Porto Alegre, Brazil\\
 \texttt{joaoschapke@gmail.com} \\

 \And
 Jonas Kubilius \\
 Three Thirds\\
 \texttt{jonas@threethirds.ai} \\

 \And
 Jurgis Pašukonis \\
 Three Thirds\\
 \texttt{jurgis@threethirds.ai} \\

 \And
 Linas Klimas \\
 Three Thirds\\
 \texttt{linas@threethirds.ai} \\

 \And
 Matthew Hausknecht \\
 Microsoft Research\\
 \texttt{matthew.hausknecht@microsoft.com} \\

 \And
 Patrick MacAlpine \\
 Microsoft Research\\
 \texttt{patmac@gmail.com} \\

 \And
 Quang Nhat Tran \\
 Temple University\\
 \texttt{quangtran@temple.edu} \\

 \And
 Thomas Tumiel \\
 AIcrowd Community\\
 \texttt{ttumiel@gmail.com} \\

 \And
 Xiaocheng Tang \\
 DiDi Labs, Mountain View, CA\\
 \texttt{xiaochengtang@didiglobal.com} \\

 \And
 Xinwei Chen \\
 AIcrowd Community\\
 \texttt{o.xlnwel@outlook.com} \\

 \And
 Christopher Hesse \\
 OpenAI\\
 \texttt{csh@openai.com} \\

 \And
 Jacob Hilton \\
 OpenAI\\
 \texttt{jhilton@openai.com} \\

 \And
 William Hebgen Guss \\
 OpenAI\\
 \texttt{wguss@openai.com} \\

 \And
 Sahika Genc \\
 Amazon Web Services Artificial Intelligence\\
 \texttt{sahika@amazon.com} \\

 \And
 John Schulman \\
 OpenAI\\
 \texttt{joschu@openai.com} \\

 \And
 Karl Cobbe \\
 OpenAI\\
 \texttt{karl@openai.com} \\
}

\begin{document}
\maketitle

\begin{abstract}

The NeurIPS 2020 Procgen Competition was designed as a centralized benchmark with clearly defined tasks for measuring Sample Efficiency and Generalization in Reinforcement Learning. Generalization remains one of the most fundamental challenges in deep reinforcement learning, and yet we do not have enough benchmarks to measure the progress of the community on Generalization in Reinforcement Learning. We present the design of a centralized benchmark for Reinforcement Learning which can help measure Sample Efficiency and Generalization in Reinforcement Learning by doing end to end evaluation of the training and rollout phases of thousands of user submitted code bases in a scalable way. We designed the benchmark on top of the already existing Procgen Benchmark by defining clear tasks and standardizing the end to end evaluation setups. The design aims to maximize the flexibility available for researchers who wish to design future iterations of such benchmarks, and yet imposes necessary practical constraints to allow for a system like this to scale. This paper presents the competition setup and the details and analysis of the top solutions identified through this setup in context of 2020 iteration of the competition at NeurIPS.

\end{abstract}



\section{Introduction}


Procgen Benchmark is a collection of 16 procedurally generated environments designed to benchmark sample efficiency and generalization in reinforcement learning \cite{cobbe2020leveraging}. Since all content is procedurally generated, each Procgen environment intrinsically requires agents to generalize to never-before-seen situations. Critical elements like level difficulty, level layout, and in-game assets are randomized at the start of every episode. These environments therefore provide a robust test of an agent’s ability to learn in many diverse settings. By aggregating performance across so many diverse environments, Procgen Benchmark provides high quality metrics to judge RL algorithms. Furthermore, Procgen environments are easy to use\footnote{All environments are open-source and can be found at \href{https://github.com/openai/Procgen}{https://github.com/openai/procgen}} and the environments are computationally lightweight. Individuals with limited computational resources can easily reproduce baseline results and run new experiments, and this ability to iterate quickly can help accelerate research. 

In this paper, we present the competition design and results of the NeurIPS 2020 Procgen Benchmark. The goal of this competition was to demonstrate the feasibility of using Procgen Benchmark to collectively measure progress on sample efficiency and generalization in Reinforcement Learning.
Prior to our competition, reinforcement learning competitions have focused on the development of policies or meta-policies which perform well on a complex domain or across a select set of tasks \cite{kidzinski2018learning, nichol2018gotta, guss2019minerl}. However, to the best of our knowledge, our competition is the first to directly isolate and focus on generalization in reinforcement learning across a broad set of procedural generated tasks. Additionally while the recent MineRL competition utilizes a procedurally generated environment, the competition is more restricted in its procedural generation engine than Procgen and focuses on algorithmic invariance to domain shift as opposed to true generalization across a task distribution \cite{guss2019minerl}.




\section{Competition}

\subsection{Environments} \label{appendix:env_disc}


This competition builds upon 16 procedurally generated environments which were publicly released as a part of the Procgen Benchmark \cite{cobbe2020leveraging}. 4 hold-out test environments were created for the evaluations of this competition and were used for the end to end evaluation of the submissions in the competition.  Although other environments such as MineRL \cite{gussminerlijcai2019},  Malmo \cite{johnson2016malmo}, and Jelly Bean World \cite{platanios2020jelly} make use of procedural generation, Procgen's novelty is in including many \emph{diverse} procedurally generated environments. Further, the Arcade Learning Environment \cite{ale} is very widely used to judge RL algorithms, but it doesn't require agents to meaningfully generalize. This is a significant flaw as generalization is critical in many real world tasks, and it's important that RL benchmarks reflect this reality.

In all environments, procedural generation controls the selection of game assets and backgrounds, though some environments include a more diverse pool of assets and backgrounds than others. When procedural generation must place entities, it generally samples from the uniform distribution over valid locations, occasionally subject to game-specific constraints. Several environments use cellular automata \cite{johnson2010cellular} to generate diverse level layouts.

\subsection{Metrics}
\label{sec:metrics}

To compare submissions based on a single score across multiple Procgen environments, we calculate the mean normalized return. Following the original Procgen paper, we define the normalized return to be $R_{norm} = (R - R_{min}) / (R_{max} - R_{min})$, where $R$ is the raw expected return and $R_{min}$ and $R_{max}$ are constants chosen (per environment) to approximately bound $R$. As each of the Procgen environments have a clear score ceiling, it was possible to establish these constants. Using this definition, the normalized return is (almost) guaranteed to fall between 0 and 1. Since Procgen environments are designed to have similar difficulties, it's unlikely that a small subset of environments will dominate this signal. We use the mean normalized return since it offers a better signal than the median, and since we do not need to be robust to outliers. 

The score computation in Round-1 uses a weighted metric on top of the mean normalized returns. When computing the cumulative round-1 score, same weight is provided to the normalized return from the evaluations of the single test environment as that is provided to the normalized return from the evaluations of all the 16 public environments. This is intentionally designed to incentivize participants to submit their training phase code while they experiment independently with the publicly released Procgen environments. 

As mentioned in Section \ref{sec:training_phase}, considerations around sample-efficiency are imposed (when necessary) by limiting the total number of timesteps during the training phase to 8M timesteps. And considerations around generalizability are imposed (when necessary) by limiting the total number of levels an agent has access to during the training phase.

\subsection{Tasks}
\label{sec:tasks_and_applications}




The competition was divided into four separate rounds : Warm-Up Round, Round-1, Round-2, Final Exhaustive Evaluation.
The evaluation of the submissions for this competition was done across two independent phases : \textbf{Training Phase} and \textbf{Rollout Phase}.

\subsubsection{Training Phase and Rollout Phase}

\subthreesection{Rollout Phase}
\label{sec:evaluation_phase}
The rollout phase focuses on evaluating a trained model (a checkpoint) against a set of Procgen environments.
The trained model used in this phase, was a carry-forward asset generated after the successful execution of the corresponding Training Phase (\ref{sec:training_phase}). 
The environments in each evaluation set were drawn from either the 16 \textbf{publicly released environments} or the 4 \textbf{hold-out test environments} created specifically for this competition. Whenever the hold-out test environments were used for the evaluation, the corresponding Training Phase (\ref{sec:training_phase}) had to be invoked by design - as the hold-out test environments were not accessible to participants in advance.

\subthreesection{Training Phase}
\label{sec:training_phase}
The training phase focuses on orchestrating the user submitted code to train against one of the Procgen environments. The training phase always generated a model checkpoint which was subsequently used in the corresponding rollout phase (\ref{sec:evaluation_phase}). The authors would like to re-iterate the fact that, whenever a hold-out test environment is used in the rollout phase, the corresponding training phase has to be invoked by design.
\\
\\
Two key things that are taken into consideration during the training phase are \textbf{sample efficiency} and \textbf{generalizability}.

As was shown by the MineRL Competition, limiting the number of environment steps in training yields resource- and sample-efficient submissions that perform well despite this limitation. We likewise address considerations around sample efficiency by limiting the number of timesteps allowed during the training phase to the same 8M timesteps \cite{guss2019minerl}.

Considerations around generalizability are addressed by limiting the number of levels (of a particular Procgen environment) that the submitted solutions have access to during the training phase, to 200 Levels.

\subsubsection{Rounds}
\subthreesection{Warm Up Round}

\begin{figure}[h]
    \centering
    \includegraphics[width=\textwidth]{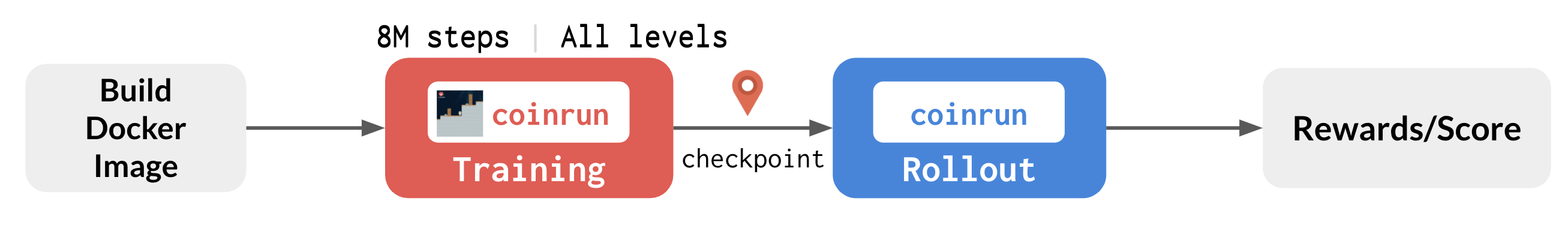}
    \caption{Evaluation workflow for the warm up round}
    \label{fig:warm-up-evaluation-setup}
\end{figure}

The goal of the Warm Up Round was to encourage participants to explore the resources (environments, starter kit, tutorials, baselines) released as a part of the competition. This round only considered only the \textit{coinrun} environment. 

The submission repository structure (included in the starter kit) required the participants to include the code for their training phase and rollout phase. Pre-agreed entrypoints for both the phases were specified in the submission repository structure. Participants were not allowed to include any trained checkpoints in their submissions. Files larger than a threshold size were automatically scrubbed from the submission repository by the AIcrowd evaluators. 

On receipt of the submissions, the AIcrowd evaluators orchestrate the submitted code for the training phase on the \textit{coinrun} environment. Considerations for Sample Efficiency are imposed by limiting the available training timesteps to 8M timesteps. Considerations for Generalization were not taken into account in this round. After the successful execution of the training phase, the trained model was carry-forwarded to the corresponding rollout phase of the submitted code, which was evaluated on the \textit{coinrun} environment on 1000 randomly sampled levels to aggregate the final scores. Figure \ref{fig:warm-up-evaluation-setup} illustrates the evaluation workflow for Warm-Up Round.

\subthreesection{Round-1}

\begin{figure}[h]
    \centering
    \includegraphics[width=\textwidth]{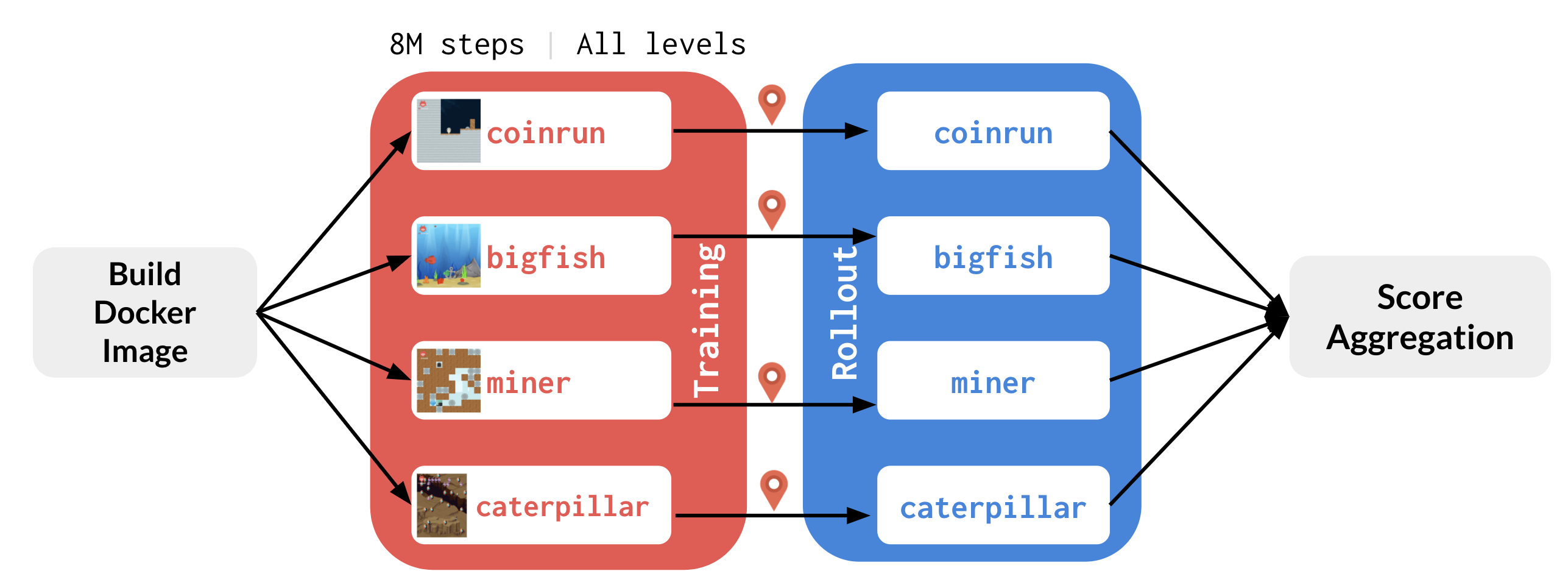}
    \caption{Evaluation workflow for the Round 1. Caterpillar is a hold-out test environment that participants did not have access to throughout the competition}
    \label{fig:round-1-evaluation-setup}
\end{figure}


Round-1 extends the problem setting of the Warm-Up Round by introducing parallel evaluations across multiple Procgen environments.  

All the submissions were evaluated against 3 public Procgen environments (\textit{coinrun}, \textit{bigifsh}, \textit{miner}) and one 1 hold-out test environment (\textit{caterpillar}), which participants did not have access to throughout the duration of the competition.

The normalized score for a single submission-environment pair was computed as described in Section \ref{sec:metrics}.

The cumulative score of each submission in this round was determined by the mean normalized score of the submission across all the 4 Procgen environments used in the Rollout phase. For reference, the normalized score for a single submission-environment pair was computed as described in Section \ref{sec:metrics}.

\begin{equation*}
    Score = \frac{1}{6} \cdot R_{norm}^{coinrun} + \frac{1}{6} \cdot R_{norm}^{bigfish} + \frac{1}{6} \cdot R_{norm}^{miner} + \frac{1}{2} \cdot R_{norm}^{caterpillar}
    \label{eq:round-1-score}
\end{equation*}

Similar to Warm-Up Round, considerations for Sample Efficiency are imposed by limiting the available training timesteps to 8M timesteps. Considerations for Generalization were not taken into account in this round.

\subthreesection{Round-2}

\begin{figure}[h]
    \centering
    \includegraphics[width=\textwidth]{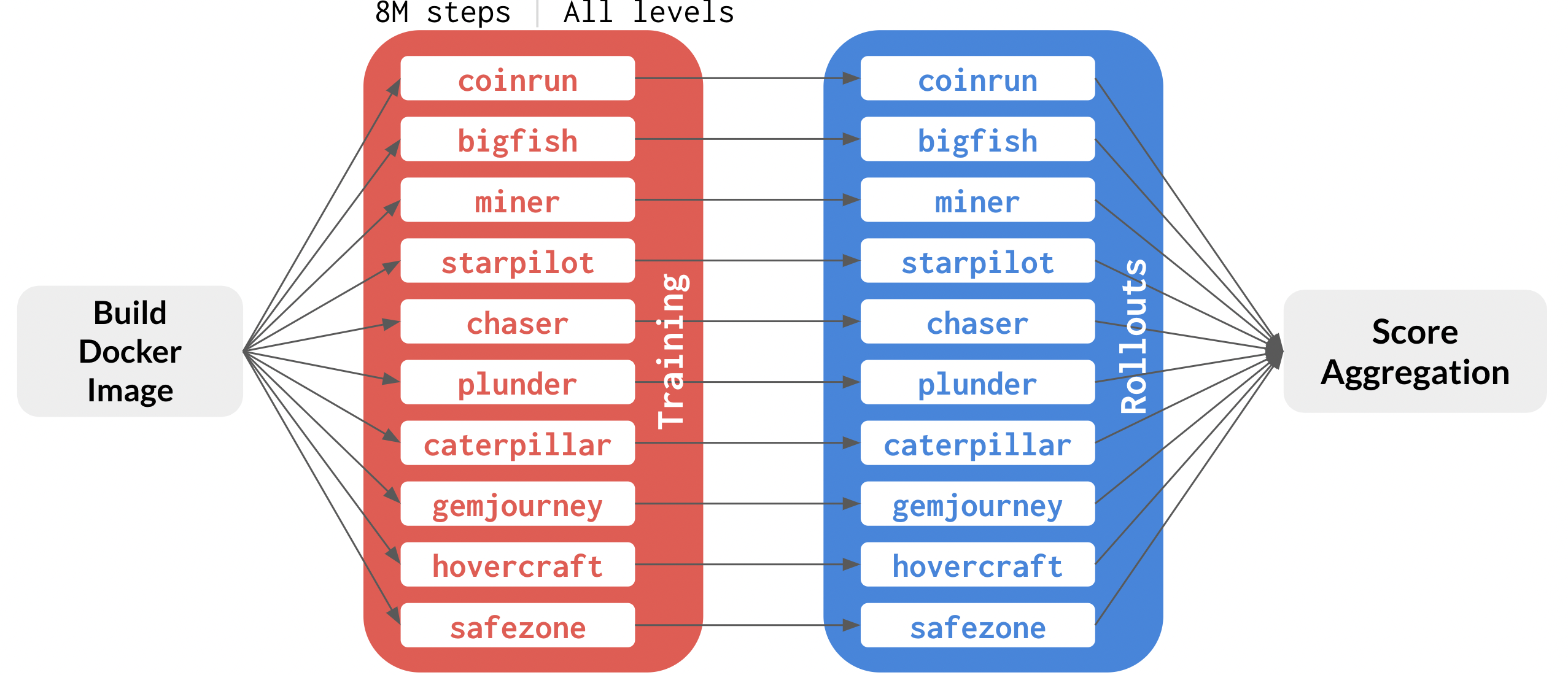}
    \caption{Evaluation setup for the Round 2. \textit{caterpillar}, \textit{gemjourney}, \textit{hovercraft}, \textit{safezone} are hold-out test environments that participants did not have access to throughout the competition.}
    \label{fig:round-2-evaluation-setup}
\end{figure}

Round-2 extends the problem setting as described in Round-1 by introducing 3 additional hold out test environments that participants did not have access to, throughout the competition. 
All the submissions in Round-2 were evaluated on 6 public environments (\textit{coinrun}, \textit{bigfish}, \textit{miner}, \textit{starpilot}, \textit{chaser}, \textit{plunder}) and 4 hold out test environments (\textit{caterpillar}, \textit{gemjourney}, \textit{hovercraft}, \textit{safezone}) for both the Training Phase and the Rollout Phase. 


This round continued to impose Sample Efficiency considerations by limiting the training phase to 8M timesteps. Considerations for Generalization were introduced in this round, where the training phase was limited to 200 levels for each of the environments, while the trained models were evaluated on 1000 randomly sampled levels during the Rollout phase.

The cumulative score of each submission in this round was determined by the mean normalized score of the submissions in the rollout phase across 6 public environments (\textit{coinrun}, \textit{bigfish}, \textit{miner}, \textit{starpilot}, \textit{chaser}, \textit{plunder}) and 4 hold out test environments (\textit{caterpillar}, \textit{gemjourney}, \textit{hovercraft}, \textit{safezone}). For reference, the normalized score for a single submission-environment pair was computed as described in Section \ref{sec:metrics}.

\begin{equation*}
    Score = \frac{1}{12} \cdot \sum_{Env}^{Public Envs} R_{norm}^{Env} + \frac{1}{8} \cdot \sum_{Env}^{Hold-out Envs} R_{norm}^{Env}
    \label{eq:round-2-score}
\end{equation*}

\subthreesection{Final Exhaustive Evaluation}

\begin{figure}
    \centering
    \includegraphics[width=\textwidth]{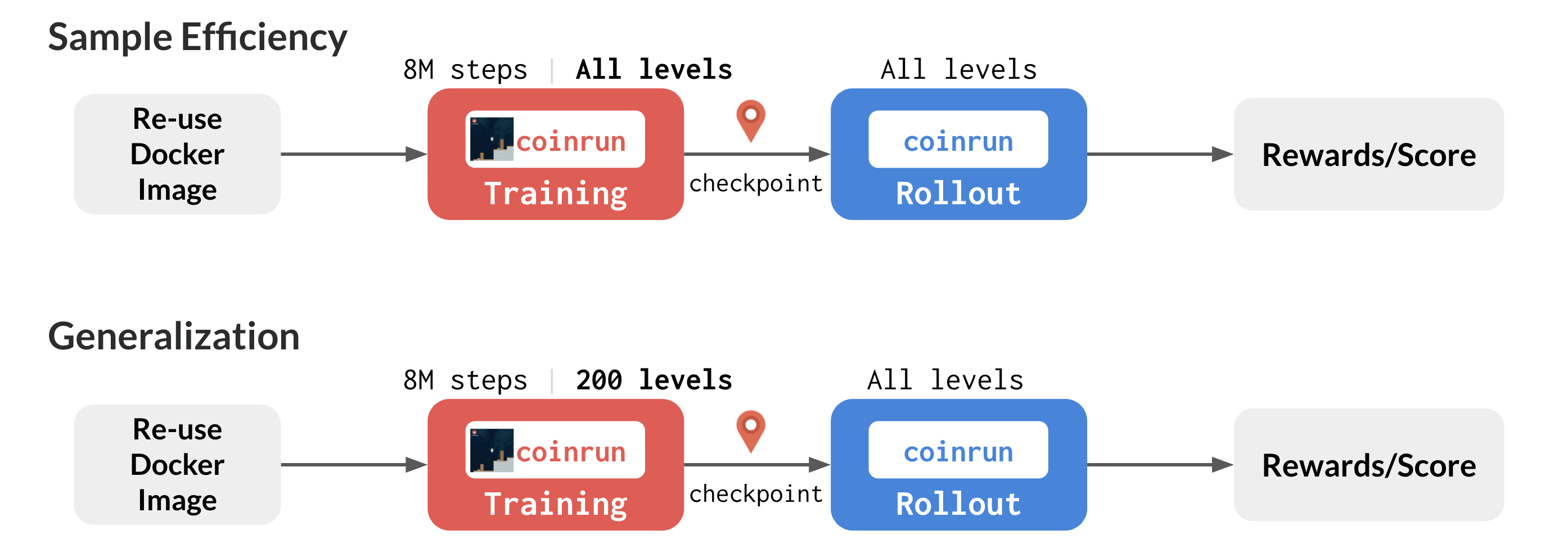}
    \caption{Evaluation workflow for sample efficiency and generalization tracks}
    \label{fig:sample-efficiency-generalization-overview}
\end{figure}

The top-10 teams from Round-2 were subject to a final exhaustive evaluation.
The final exhaustive evaluation is an extended and robust version of the problem formulation of Round-2. 

The top-10 teams had the option to specify separate submissions for both the Sample Efficiency and Generalization tracks. 

All eligible submissions were evaluated on all the 16 public environments and 4 hold out test environments for both the Training Phase and the Rollout Phase. The submissions were evaluated separately for Sample Efficiency and for Generalization. When measuring Generalization, the constraints of 8M timesteps from Sample Efficiency were implicitly added for operational reasons. Considerations around Generalization were imposed by limiting the number of levels during the training phase to 200 levels for each of the 20 procgen environments. During the rollouts phase the trained models for each environment were evaluated on 1000 randomly sampled levels.

All eligible submissions were evaluated over 3 trials for the Sample Efficiency Track and 3 trials for the Generalization Track. In the Sample Efficiency track, no limits on the number of levels were imposed during the training phase, while in the Generalization track limited the number of levels during the training phase to 200 levels. 

The score for each trial was determined by the mean normalized score of the submissions in the rollout phase across all the 20 procgen environments. 
\begin{equation*}
    score_{trial_n} = \frac{1}{20} \cdot \sum_{Env}^{All\ Envs} R_{norm}^{Env}
    \label{eq:final-evaluation-score}
\end{equation*}

The final score for each of the tracks were determined by the maximum score across the three task specific trials for both the Sample Efficiency and the Generalization track.

\begin{equation*}
    score_{generalization} = \max_{1 \leq n \leq 3} \{{score^{generalization}_{trail_n}}\} 
    \label{eq:final-score-genralization}
\end{equation*}
\begin{equation*}
    score_{sample\_efficiency} = \max_{1 \leq n \leq 3} \{{score^{sample\_efficiency}_{trail_n}}\}
    \label{eq:final-score-sample-efficiency}
\end{equation*}

\subsection{Competition Statistics}
Throughout the competition, a total of 545 individual participants, spread across 44 countries, registered for the competition. After round 1, 50 teams (83 individual participants) were eligible to participate in the round 2. Finally, 10 teams containing 18 individual participants qualified for the final exhaustive evaluation. Across the whole competition, we evaluated a total of 4805 submissions resulting in over 172,000 trained checkpoints throughout the competition and across different evaluation configurations.

\section{Methods}

\begin{figure}[htbp]

  \label{fig:SampleEfficiency}%
  {\caption{The mean normalized rewards across training per environment for top ten teams in sample-efficiency track.}}%
  {\includegraphics[width=0.99\linewidth]{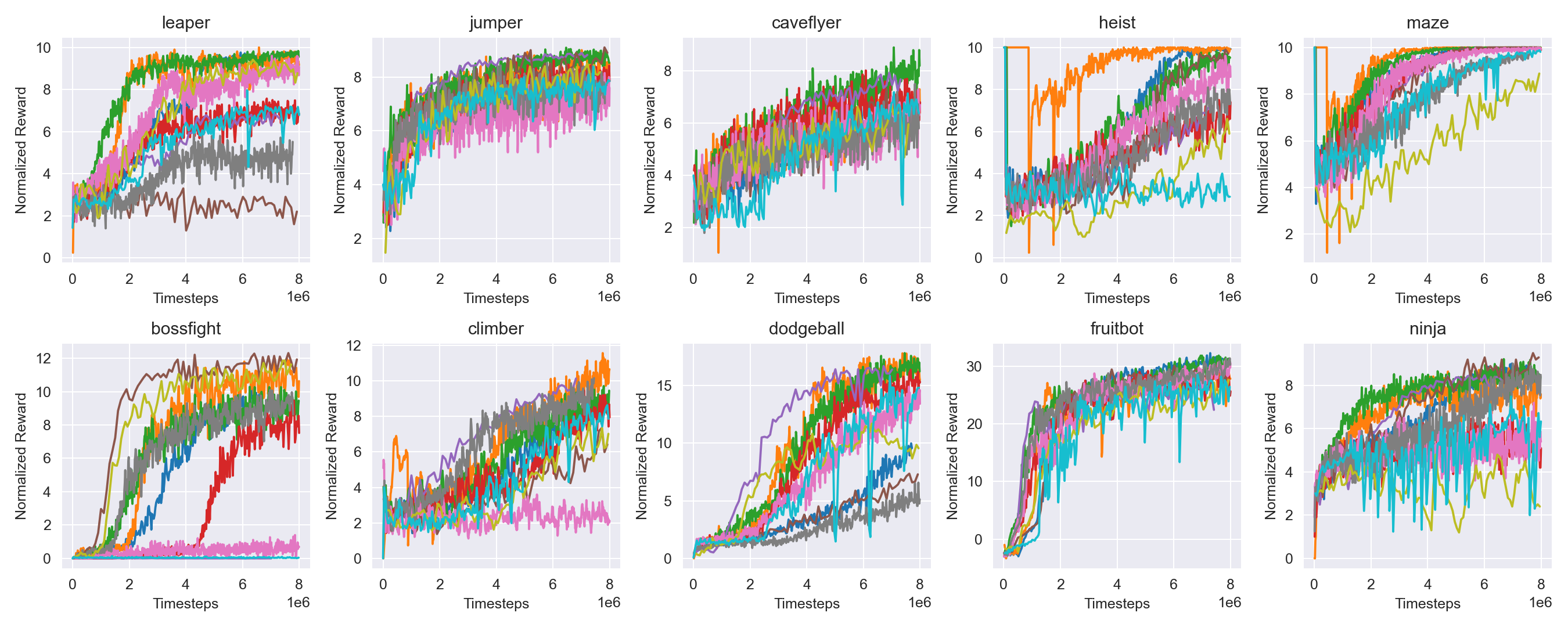}}
\end{figure}

\begin{figure}[htbp]

  \label{fig:Generalization}%
  {\caption{The mean normalized rewards during training phase across rollouts per environment for Team MSRL in generalization track. Note that training score are for the 200 levels, not for the entire distribution of the environment.}}%
  {\includegraphics[width=0.95\linewidth]{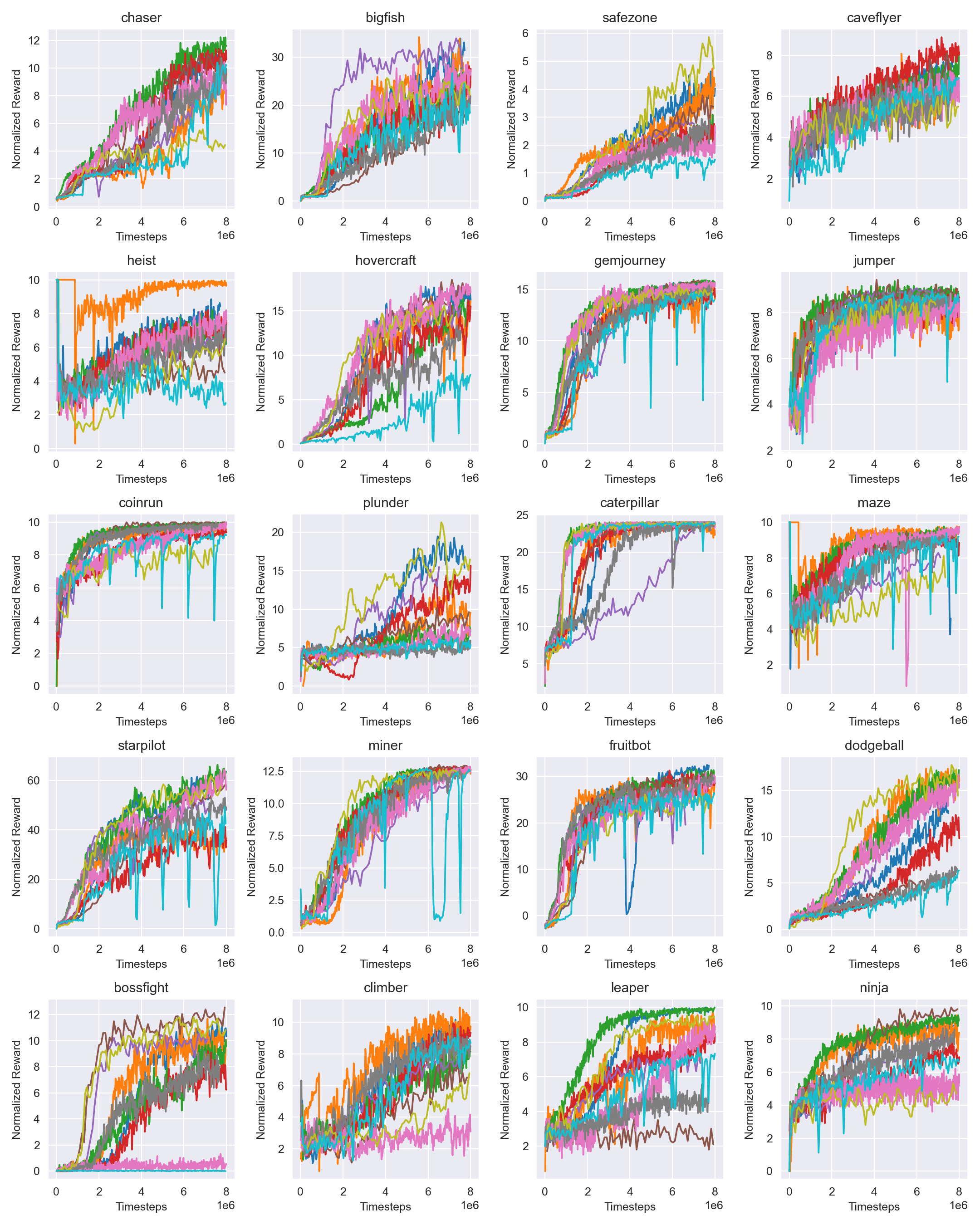}}
\end{figure}

The methods used by the top-10 teams can be broadly cateogorized into a set of ``base algorithms``. The list of ``base algorithms`` used by the participants is shown in Table~\ref{tab:ListofMethods}.
The competitors were limited to using a single NVIDIA V100 GPU and 2 hours of training for their models with 8M steps. We used the mean normalized return to compare submissions based on a single score across multiple Procgen environments. 
The mean normalized rewards across multiple rollouts per environment for the top ten submission for sample-efficiency and generalization tracks are shown in \ref{fig:SampleEfficiency} and \ref{fig:Generalization}. In both tracks, the trends for the mean normalized rewards for rollouts were similar for \textit{jumper, caveflyer, maze} and \textit{fruitbot} for most of the top ten submissions while the final normalized rewards varied significantly in \textit{leaper} and \textit{dodgeball} for several competitors.

In the following, we provide the key implementation details and modifications for each of the top-10 teams. We grouped the participants' methods into three categories based on the base algorithm: 1) Phasic Policy Gradient (PPG) \cite{cobbe2020phasic}, 2) Proximal Policy Optimization (PPO) \cite{ppo}, and 3) other, including Policy-on Policy-off Policy Optimization (P3O) \cite{p3o}, Reactor \cite{gruslys2017reactor}, and Soft Actor-Critic (SAC) \cite{haarnoja2018soft}. 

\begin{table}[htbp] \label{tab:ListofMethods}
\centering
\caption{Ranking of participants for generalization and sample efficiency and their base algorithms. The Baseline provided by the organizers replicated the PPO implementation as described in \cite{cobbe2020leveraging}}
\begin{tabular}{|p{2cm}|p{1.5cm}|p{1.5cm}|p{1.75cm}|p{1.75cm}|p{5cm}|}
\hline
\textbf{Team or Individual} & \textbf{General. Rank} & \textbf{Sample Eff. Rank} & \textbf{General. Mean Norm. Reward} & \textbf{Sample Eff. Mean Norm. Reward} & \textbf{Base Algorithm}                                     \\ \hline
TRI & 1  & 1 & 0.6083 &	0.7680  & Phasic Policy Gradient      \\ \hline
MSRL & 2 & 7 &0.5290 & 0.6700 & Proximal Policy Opt.        \\ \hline
Alpha & 3  & 4 & 0.5193 & 0.7071 & -                          \\ \hline

ttom & 4 & 9 & 0.4939 &	0.6386 & Proximal Policy Opt.        \\ \hline
Gamma & 5 & 3 & 0.4898 & 0.7231 & Phasic Policy Gradient    \\ \hline
zero & 6 & 8 & 0.4699 &	0.6431  & Soft Actor-Critic         \\ \hline
Xiaocheng Tang  & 7 & 5 & 0.4523 & 0.6918 & Proximal Policy Opt.\\ \hline
Joao Schapke & 8 & 2 & 0.4447 & 0.7342  & Policy-on Policy-off Policy Opt. \\ \hline
Paseul & 9 & 10 & 0.3963 & 0.5847 &  -  \\ \hline
three\_thirds & 10  & 6 &0.3694 & 0.6916 & Reactor \& Soft-Actor Critic \\ \hline

\textbf{Baseline}     &  \textbf{11}                      & \textbf{11}    & \textbf{0.2002} &  \textbf{0.3695} &    \textbf{Proximal Policy Opt.} \\ \hline
\end{tabular}
\end{table}

\subsection{Base Algorithm: Phasic Policy Gradient}
The top two teams in generalization category used variations of PPG \cite{cobbe2020phasic}. Their PPG-based methods also improved sample-efficiency, ranking both teams among the top three. PPG is an extension of PPO that, among other improvements, allows for greater sample reuse by (partially) separating policy and value-function learning into separate phases \cite{cobbe2020phasic}.

\subsubsection{Team: Gamma} 

We used the value function from the latest PPO iteration is better instead of the older value targets in the auxiliary phase. We discovered that recalculating the value targets on the entire replay buffer using GAE \cite{Schulmanetal_ICLR2016}  boosts sample efficiency. We augment the observations in the replay buffer during the auxiliary phase. Crucially, we found that keeping a decent percentage of frames un-augmented is important for policy stability. We used random translate, and colored cutout augmentations from \cite{laskin2020reinforcement} and apply them consistently across frame stacks. Using highest probability action during inference led to the agent getting stuck in one place when a wrong action led to the same state. However, we thought that the stochastically choosing actions in unseen environments led to too much randomness and premature death. Thus, we reduced the softmax temperature during inference which led to improvements in the scores for all environments. Hyperparameter tuning played a very crucial role in the competition. The generalization and sample-efficiency results of the final submission are shown in \Cref{fig:GammaGeneralization} and \Cref{fig:GammaSampleEfficiency} respectively.

\subsubsection{Team: TRI}

For computational reasons, we elected to use the single-network variant of PPG, though unlike the original paper we left the value head of the network attached during the policy phase, and used a smaller loss coefficient for the value objective. We adapted PPO and PPG to use data augmentation. We evaluated a subset of the augmentations such as translation, vertical and horizontal flipping, rotation, conversion to grayscale, and color cutout, finding that random translation worked most consistently across environments. We found that augmentation during the auxiliary phase worked best, and used that. We used a reward shaping penalty. Reward normalization, which involves transforming the rewards of the agent with the goal of normalizing the learning targets of the value network, had a significant impact on performance. Finally, we performed coarse grid searches for a subset of hyperparameters. 


The generalization and sample-efficiency results of the final submission are shown in \Cref{fig:TriGeneralization} and \Cref{fig:wulfebwSampleEfficiency} respectively.

\subsection{Base Algorithm: Proximal Policy Optimization}
One third of the top ten winners used PPO \cite{ppo} with modifications to the original IMPALA network along with variations on exploration and regularization. All the participants used some form of hyperparameter tuning.

\subsubsection{Team:MSRL}

Our approach focused on improving the performance of the basic PPO with the IMPALA agent architecture, as described in the original Procgen paper \cite{cobbe2020leveraging}, purely using data augmentation, L2 regularization of the agent network parameters, and hyperparameter tuning. We didn't find data augmentation, as well as batchnorm and dropout, to be helpful in improving performance. However, L2 regularization and careful hyperparameter choice improved PPO's performance significantly. They allowed our agent to vastly outperform the competition's PPO baseline and to approach the performance of more sophisticated techniques without any changes to the basic PPO algorithm. We tuned all hyperparameters of the rllib's PPO implementation, the number and size of the IMPALA architecture's residual blocks, and the L2 regularization coefficient via a series of hyperparameter searches using the Distributed Grid Descent algorithm \cite{loynd2020working} running on Microsoft Azure Batch service \cite{AZB}. The best-performing parameter combination we discovered is in Appendix \ref{sec:msrl}. Tuning the discount factor $\gamma$, GAE $\lambda$, L2 regularization coefficient, and the learning rate for a given architecture was particularly important. The generalization and sample-efficiency results of the final submission are shown in \Cref{fig:MSRLGeneralization} and \Cref{fig:MSRLSampleEfficiency} respectively.

\subsubsection{Individual:ttom}

The biggest single improvement to performance came from modifying the model. The baseline IMPALA CNN used residual blocks with 16, 32, and 32 channels each. Increasing this width to 32, 64, and 64 channels drastically improved performance. Average pooling the final convolutional layer before flattening and passing into a fully connected layer also improved performance and drastically reduced the number of network parameters. A good learning rate and an approximate one-cycle \cite{smith2018superconvergence} schedule greatly sped up early training, getting the algorithm to a reasonable level of performance (close to the final level) after 4M timesteps for all environments. Each algorithm and model variation had parameters tuned individually. Adding an intrinsic reward signal (Random Network Distillation, \cite{burda2018exploration}) did result in marginally faster training but performance remained approximately the same. The implementation of RND did not apply across episode boundaries, as originally implemented, which may have negatively affected the results. 
The generalization and sample-efficiency results of the final submission are shown in \Cref{fig:ttomGeneralization} and \Cref{fig:ttomSampleEfficiency} respectively.

\subsubsection{Individual:Xiaocheng}

We used Random Network Distillation (RND) bonus to encourage exploration. We observed better performance when the intrinsic reward is treated as a life-long novelty computed in a non-episodic setting regardless of the episodic nature of the tasks. Similar to the Never-Give-Up (NGU) agent, we used separate parameterizations to learn varied degrees of exploration and exploitation policies, such that the exploitative policy focuses on maximizing the extrinsic reward and the exploratory ones seek for novelty bonus. We implemented this family of policies under the PPO framework. We apply the mixup regularization proposed for supervised learning in the context of reinforcement learning to increase data diversity and to induce a smoother policy which can generalize better. In particular, we augment the training data by sampling new observations from the convex hull of distinct observations and the corresponding training targets. We added an auxiliary loss term. The auxiliary loss for training the value function can be derived in the similar fashion. To adaptively adjust the agent's behavior accordingly, we employed a meta-controller implemented as a multi-arm UCB bandit with $\epsilon_{UCB}$-greedy exploration to adapt to the changes of the reward through time. The generalization and sample-efficiency results of the final submission are shown in \Cref{fig:XiaochengGeneralization} and \Cref{fig:XiaochengSampleEfficiency} respectively.

\subsection{Base Algorithm:  Policy-on Policy-off Policy, Reactor, or Soft Actor-Critic}

A modified version of the the Policy-on Policy-off Policy Optimization (P3O) \cite{p3o} algorithm ranked second in sample-efficiency. P3O has an off-policy optimization phase in which samples are used with an policy gradient objective modulated by a clipped importance sampling term, hence, resulting in a more sample-efficient approach. The two other base algorithms used were Reactor \cite{gruslys2017reactor} and Soft Actor-Critic (SAC) \cite{haarnoja2018soft}.

\subsubsection{Team:Three-thirds}

Reactor \cite{gruslys2017reactor} combines the best of PPO and Rainbow and improves on them: it is an Actor-Critic architecture that uses a multi-step off-policy Q-learning, distributional Q-learning in the critic, and a prioritized experience replay for sequences. We modified the vanilla Reactor slightly. Instead of using LSTM, we stacked the last two images. We passed the total reward the agent had accumulated in the current episode as an additional input channel to provide additional information about the game state to the initial convolutional layers. We modified the IMPALA encoder, $[24, 40, 48]$ channels in the convolutional layers, tuned for the competition resource constraints. We used a second Q-network head as in SAC \cite{haarnoja2018soft}, which helped decrease the overestimation of the Q-value. Unlike SAC, both Q-networks shared the encoder due to limited computational resources. Sampling from the experience buffer using a mixture of prioritized and uniform distributions as in Reactor, because vanilla prioritized sampling turned out unstable. We used dynamic rescaling of the entropy coefficient $\tau$ as in SAC so that the desired entropy level is maintained. The generalization and sample-efficiency results of the final submission are shown in \Cref{fig:three_thirdsGeneralization} and \Cref{fig:three_thirdsSampleEfficiency} respectively.

\subsubsection{Individual:Joao Schapke}

P3O is a synchronous algorithm and its two phases increases the compute time of standard policy gradients. Due to the time constraints of the competition this was a limitation to the performance of the algorithm. In order to improve the compute efficiency we implemented a distributed adaptation of the original algorithm inspired by the IMPALA \cite{impala} algorithm: many actors with a single learner, and additionally, a single replay buffer. In the final distributed approach we removed the on-policy phase of the algorithm, making it an off-policy policy gradient algorithm. Used prioritised sampling to draw samples with a strong signal. The Procgen environment, similarly to Atari, does not have the Markov property. The partial observation obtained of the environment does not contain all the information needed to make optimal actions, and past observations may contain information (e.g. speed and direction of projectiles or opponents) needed for good performance. We implemented an approach that squishes past frames into a single frame. This is done by adding the weighted 1-lag difference of the past grayscaled frames. 
The generalization and sample-efficiency results of the final submission are shown in \Cref{fig:JoaoSchapkeGeneralization} and \Cref{fig:JoaoSchapkeSampleEfficiency} respectively.

\subsubsection{Team: Zero}

First, we adapted SAC for discrete action space and refine it with several improvements on DQN \cite{Mnih2015}. We then introduced a new convolutional neural network that not only performed better than the network from IMPALA \cite{Espeholt2018} but used fewer resources. We observed that each environment contains some redundant actions never used in practice. We used a shared network and allow only gradients from the $Q$ function back propagate to the encoder which worked best. We used IQN \cite{Dabney2018}, distributional RL algorithm that learns a mapping from probabilities to returns, to replace the $Q$ function in SAC. We used adaptive multi-step improves multi-step learning to allow adaptively select the multi-step target based on the quality of the experiences. We modified IMPALA CNN by adding a channel-wise module \cite{Hu2020,Woo2018} to the residual block and replacing all MaxPool with MaxBlurPool \cite{Zhang2019}. This architecture had only a quarter of the parameters compared to IMPALA CNN and empirically performed much better because of the channel-wise attention module. 
The generalization and sample-efficiency results of the final submission are shown in \Cref{fig:zeroGeneralization} and \Cref{fig:zeroSampleEfficiency} respectively.

\section{Discussion}
There were several key commonalities across the methods which resulted in significant improvements in generalization or sample-efficiency:
\begin{itemize}
\item \textit{Hyper-parameter tuning}: Almost all participants applied hyper-parameter tuning which resulted in significant improvements in the performance. The methods require a lot of tuning in order to get good performance, and this consumes a great deal of time and is a dull task. A more critical evaluation of improving hyper-parameter tuning for reinforcement learning algorithms would be beneficial for practitioners.

\item \textit{Data augmentation}: Many participants used data augmentation for generalization while trying to balance the variations without degrading the sample efficiency. For PPG, keeping a decent percentage of frames un-augmented was important for policy stability.  

\item \textit{Neural Network}: Modifying the IMPALA neural network from \cite{cobbe2020leveraging} also resulted in significant performance changes. Several teams made the IMPALA neural network's CNN layers wider. Another modification was to add channel-wise module to the residual block and replacing all MaxPool with MaxBlurPool which resulted in fewer parameters yet achieved higher rewards.

\item \textit{Reward shaping and normalization}: One team found out that the reward normalization, which involves transforming the rewards of the agent with the goal of normalizing the learning targets of the value network, had a significant impact on performance.
\end{itemize}

We also list a few methods that have not resulted in anticipated performance improvements:
\begin{itemize}
\item Adding an intrinsic reward signal 
\cite{burda2018exploration} did result in marginally faster training but performance remained approximately the same. 

\item Not all data augmentation techniques worked as well and not for all methods. In some cases, after optimizing for the neural network and algorithm parameters, the performance improvement from data augmentation was not significant.

\item Using recurrent neural networks slowed the performance too much. A more effective method was to use framestacking.

\item DQN worked on some environments but required a lot more experience and resulted in very slow training (e.g., large target network update interval, small learning rate).

\item Noisy nets \cite{fortunato2019noisy} for exploration did not result in improvements. One reason may be that the PPO update implicitly constrains the policy to not change to much, the additional change due to the noise makes the policy update less efficient.

\item Approaches based on auxiliary tasks such as contrastive learning 
\cite{srinivas2020curl} and deepMDP 
\cite{gelada2019deepmdp} degraded performance. This may be due the overhead these methods introduce, which significantly reduces the number of training steps, leading to an undertrained model.

\end{itemize}

\section{Conclusion}
We ran the NeurIPS 2021 Procgen Competition for measuring Sample Efficiency and Generalization in Reinforcement Learning. We describe the design of a Reinforcement Learning competition using the Procgen Benchmark, which enables end to end training and evaluation of thousands of user submitted code repositories. The end to end training and evaluation framework allows us to add interesting layers of complexities to the benchmark design, like the ability to enforce and measure generalization and the ability to enforce sample efficiency constraints. We summarized the performance and described the submissions of the top teams for both the Sample Efficiency Track and the Generalization Track. 

\section*{Acknowledgement}

We thank the whole team at AIcrowd for the countless hours dedicated towards the conception and execution of this challenge. We would particularly thank the whole DevOps and Support team at AIcrowd to allow such a smooth execution of such an ambitious challenge. We thank Sunil Mallya and Cameron Peron for their support in obtaining the necessary computational resources to execute a challenge of this scale, and for facilitating the provision of AWS compute credits for the participants of this benchmark. We thank Anna Luo, Jonathan Chung and Yunzhe Tao from the Amazon Sagemaker team for their continued support in preparing high quality baselines for participants to get started with Procgen Competition on Sagemaker. We thank the hundreds of participants of this competitions who spent countless hours helping us improve this benchmark and push the state of art for Generalization in RL. We thank Amazon Amazon Web Services for their generous sponsorship in computational resources to enable this competition.


\bibliographystyle{unsrt}  
\bibliography{references}  

\newpage
\appendix

\section{Supplementary Material}\label{apd:supp}
\subsection{Generalization Track Graphs}
\begin{figure}[htbp]

  \label{fig:GammaGeneralization}%
  {\caption{The mean normalized rewards during training phase across rollouts per environment for Team Gamma in generalization track. Note that training score are for the 200 levels, not for the entire distribution of the environment.}}%
  {\includegraphics[width=\textwidth]{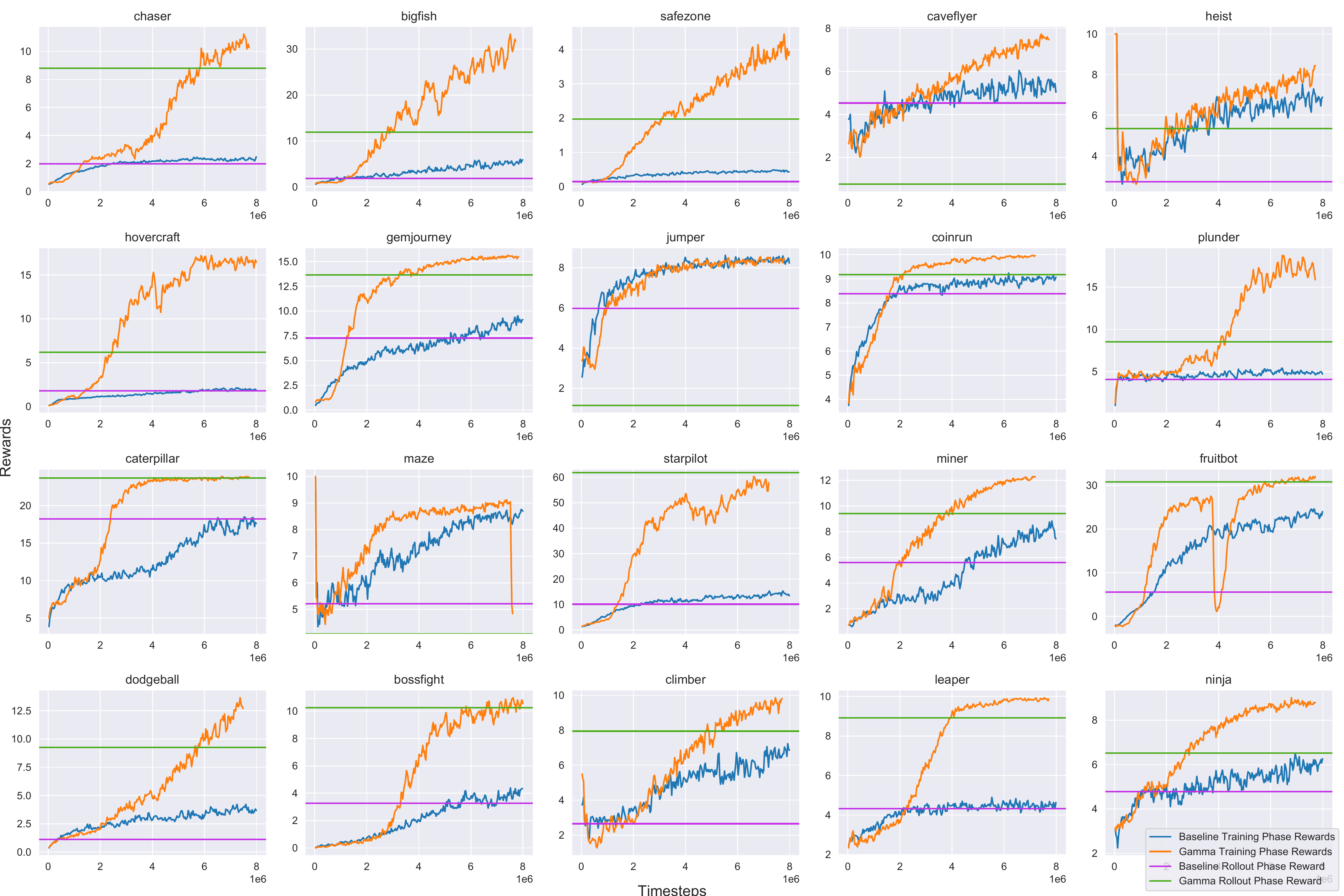}}
\end{figure}

\begin{figure}[htbp]

   \label{fig:TriGeneralization}%
  {\caption{The mean normalized rewards during training phase across rollouts per environment for Team TRI in generalization track. Note that training score are for the 200 levels, not for the entire distribution of the environment.}}%
  {\includegraphics[width=\textwidth]{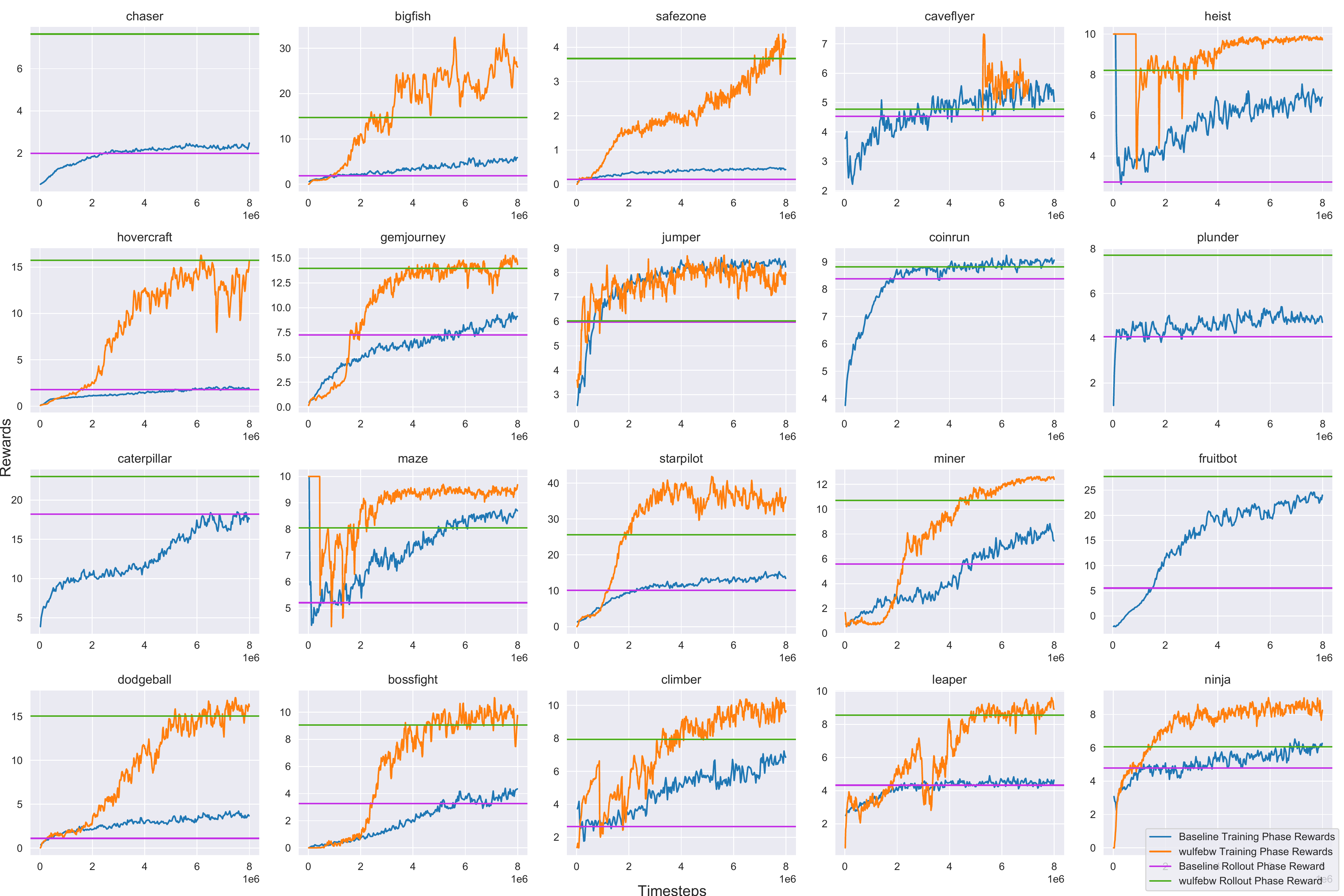}}
\end{figure}

\begin{figure}[htbp]

    \label{fig:AlphaGeneralization}%
  {\caption{The mean normalized rewards during training phase across rollouts per environment for Team Alpha in generalization track. Note that training score are for the 200 levels, not for the entire distribution of the environment.}}%
  {\includegraphics[width=\textwidth]{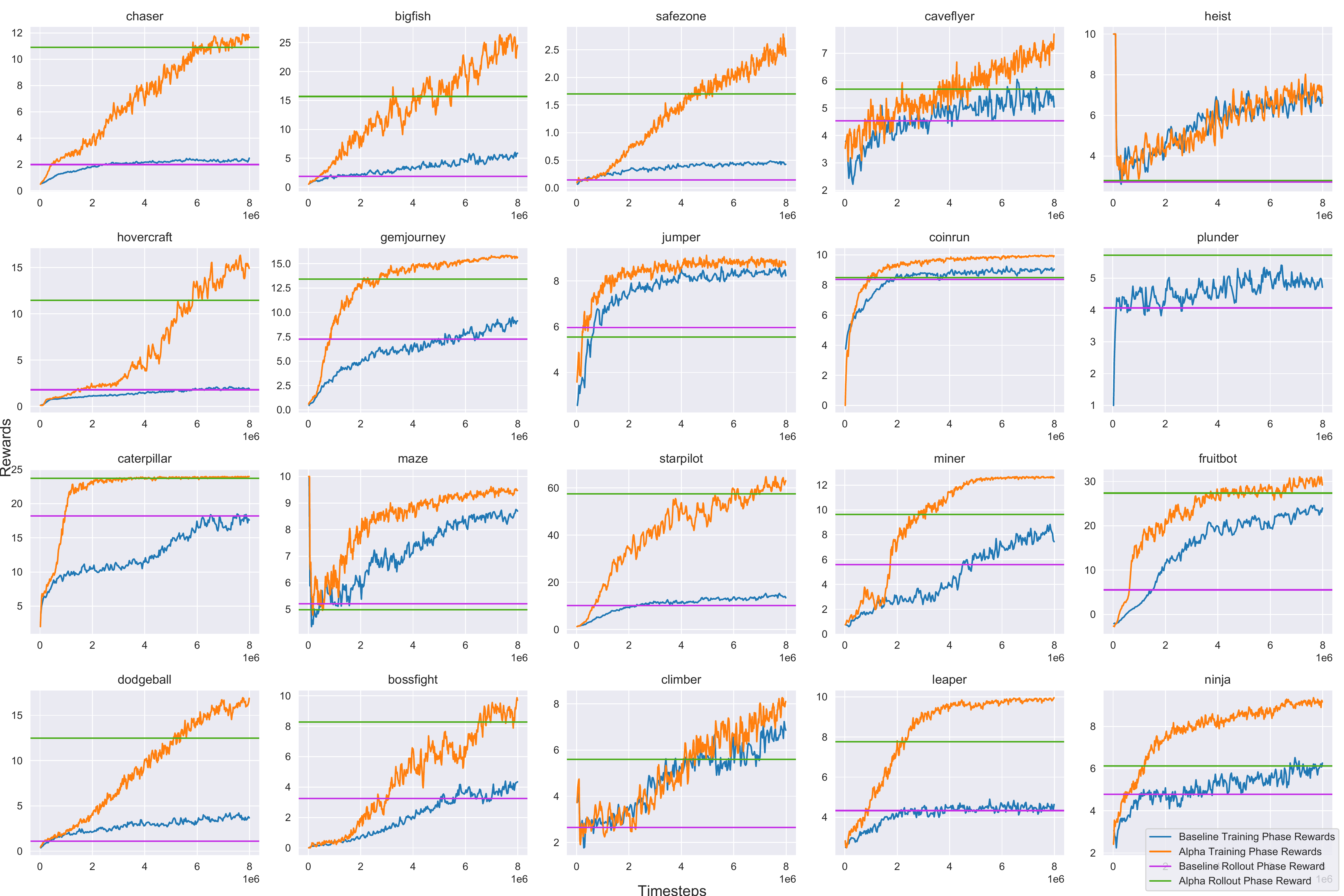}}
\end{figure}

\begin{figure}[htbp]

    \label{fig:MSRLGeneralization}%
  {\caption{The mean normalized rewards during training phase across rollouts per environment for Team MSRL in generalization track. Note that training score are for the 200 levels, not for the entire distribution of the environment.}}%
  {\includegraphics[width=\textwidth]{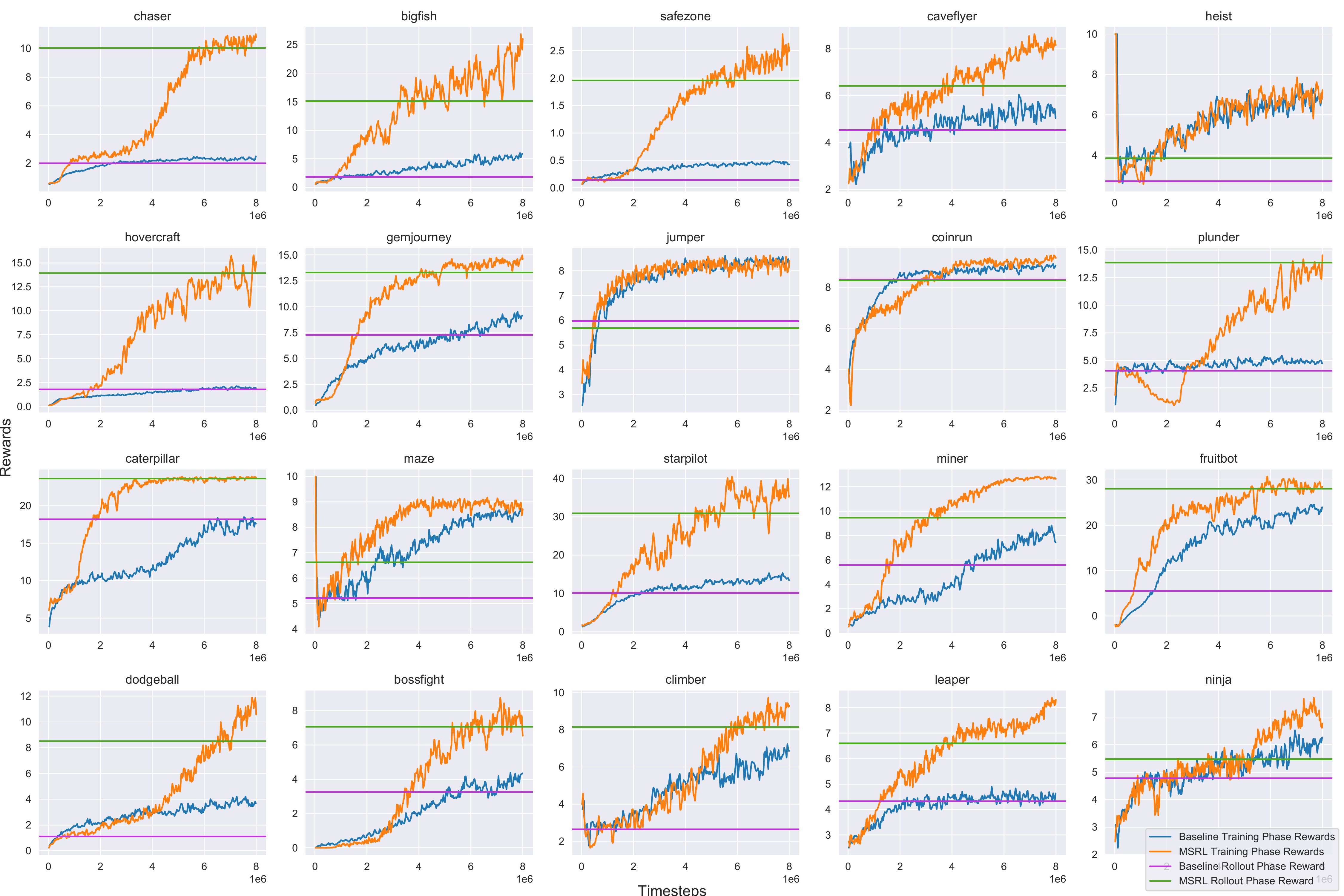}}
\end{figure}

\begin{figure}[htbp]

    \label{fig:three_thirdsGeneralization}%
  {\caption{The mean normalized rewards during training phase across rollouts per environment for Team ThreeThirds in generalization track. Note that training score are for the 200 levels, not for the entire distribution of the environment. }}%
  {\includegraphics[width=\textwidth]{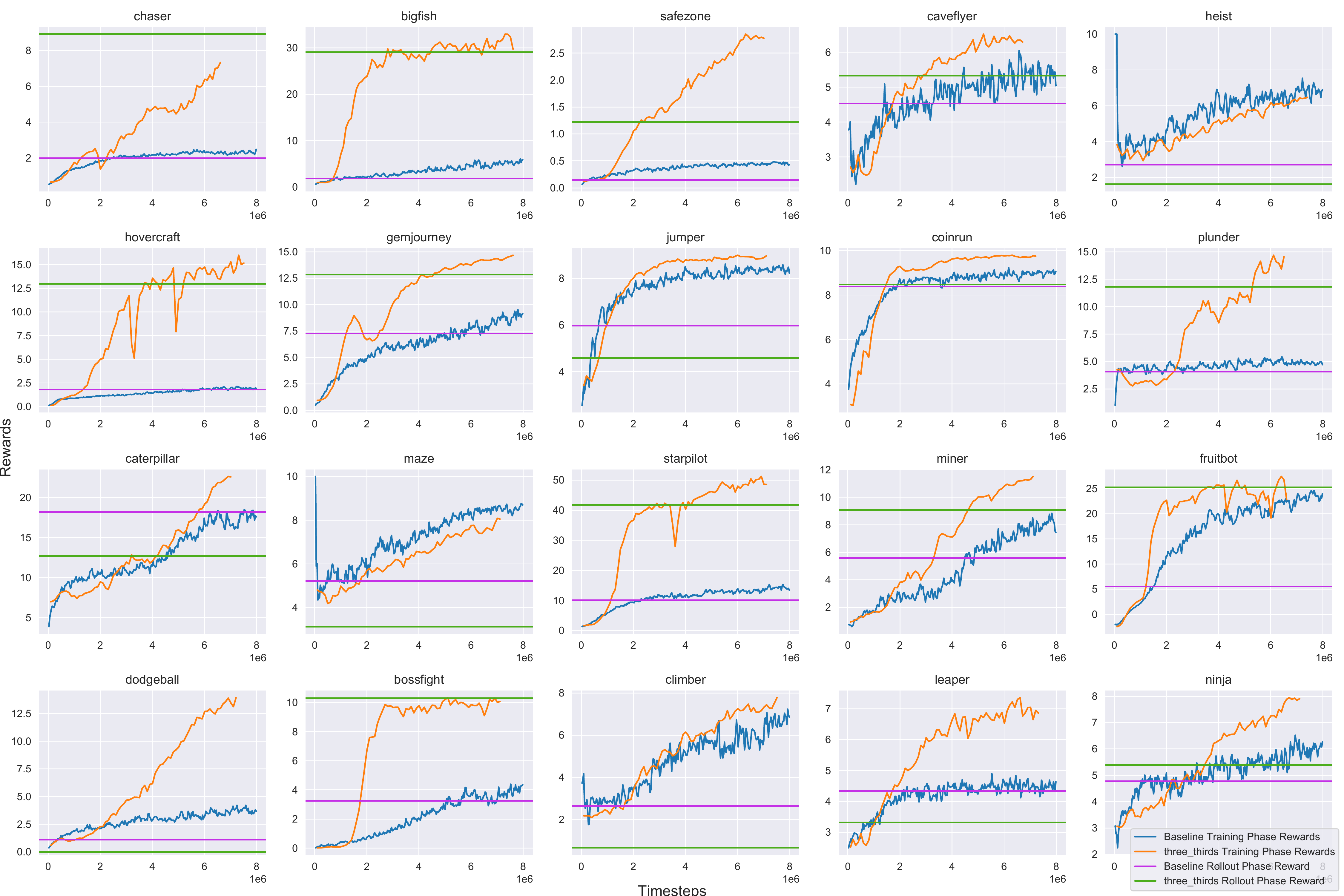}}
\end{figure}

\begin{figure}[htbp]

    \label{fig:JoaoSchapkeGeneralization}%
  {\caption{The mean normalized rewards during training phase across rollouts per environment for Individual Joao Schapke in generalization track. Note that training score are for the 200 levels, not for the entire distribution of the environment.}}%
  {\includegraphics[width=\textwidth]{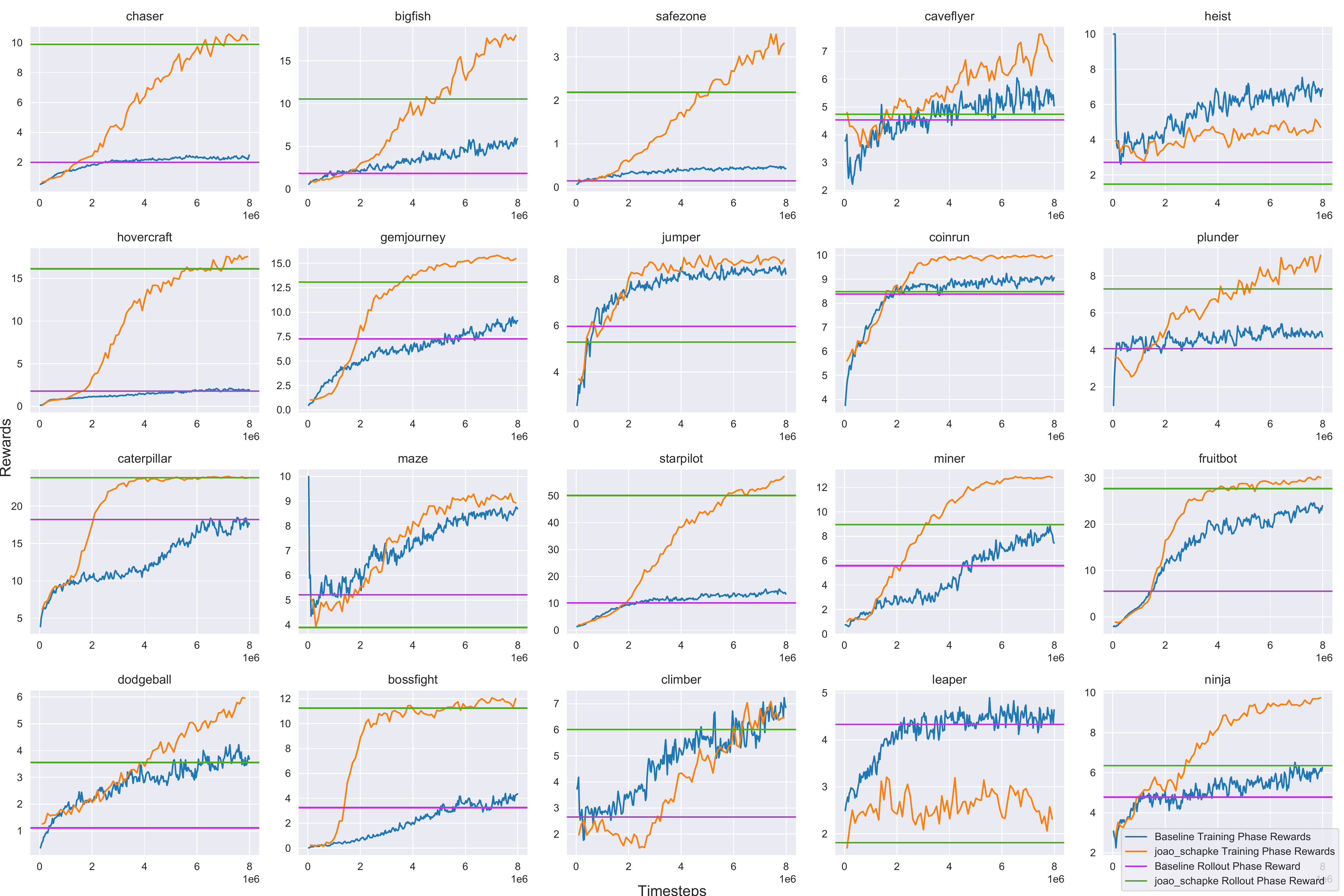}}
\end{figure}

\begin{figure}[htbp]

    \label{fig:ttomGeneralization}%
  {\caption{The mean normalized rewards during training phase across rollouts per environment for Individual ttom in generalization track. Note that training score are for the 200 levels, not for the entire distribution of the environment.}}%
  {\includegraphics[width=\textwidth]{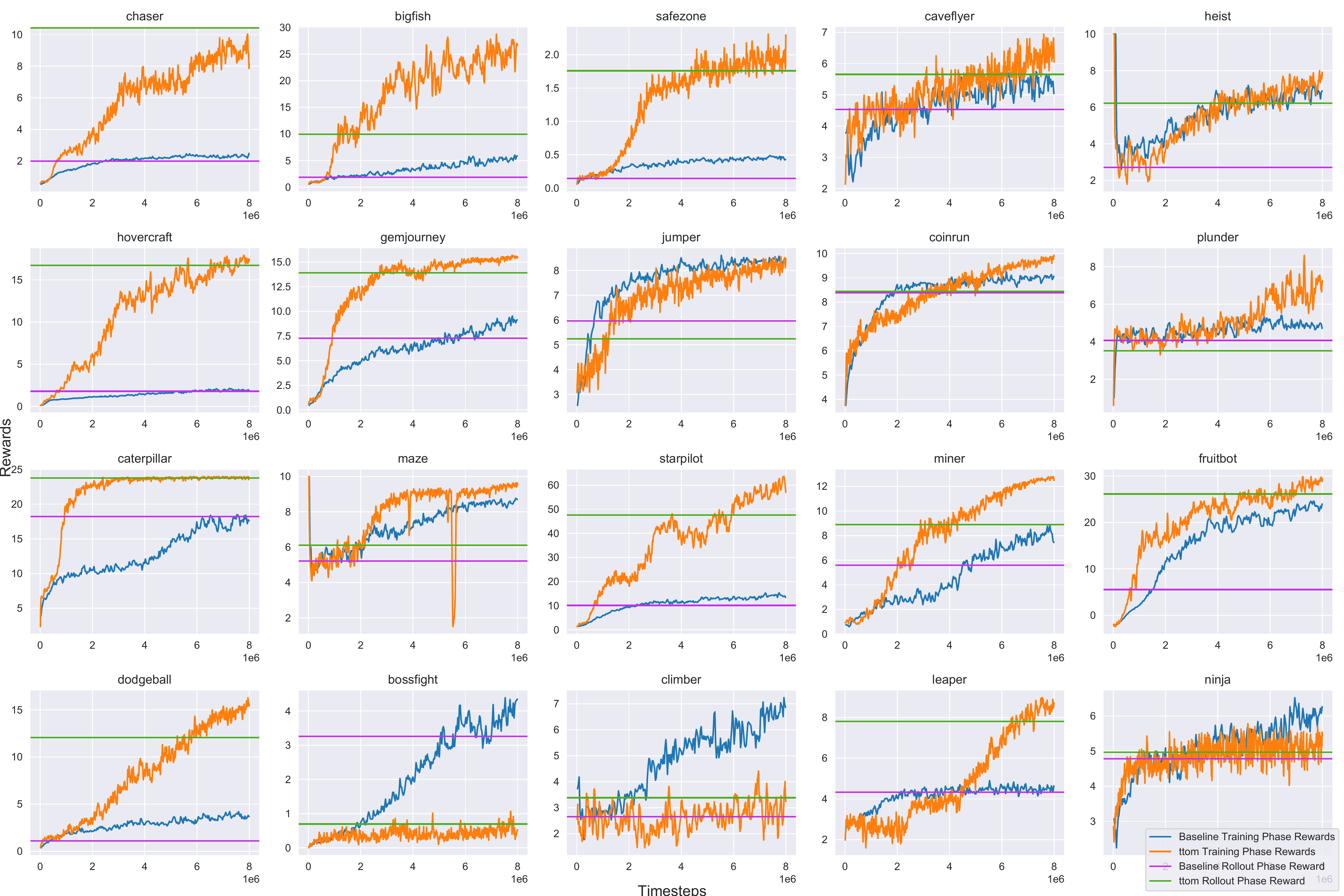}}
\end{figure}

\begin{figure}[htbp]

    \label{fig:XiaochengGeneralization}%
  {\caption{The mean normalized rewards during training phase across rollouts per environment for Individual Xiaocheng Tang in generalization track. Note that training score are for the 200 levels, not for the entire distribution of the environment.}}%
  {\includegraphics[width=\textwidth]{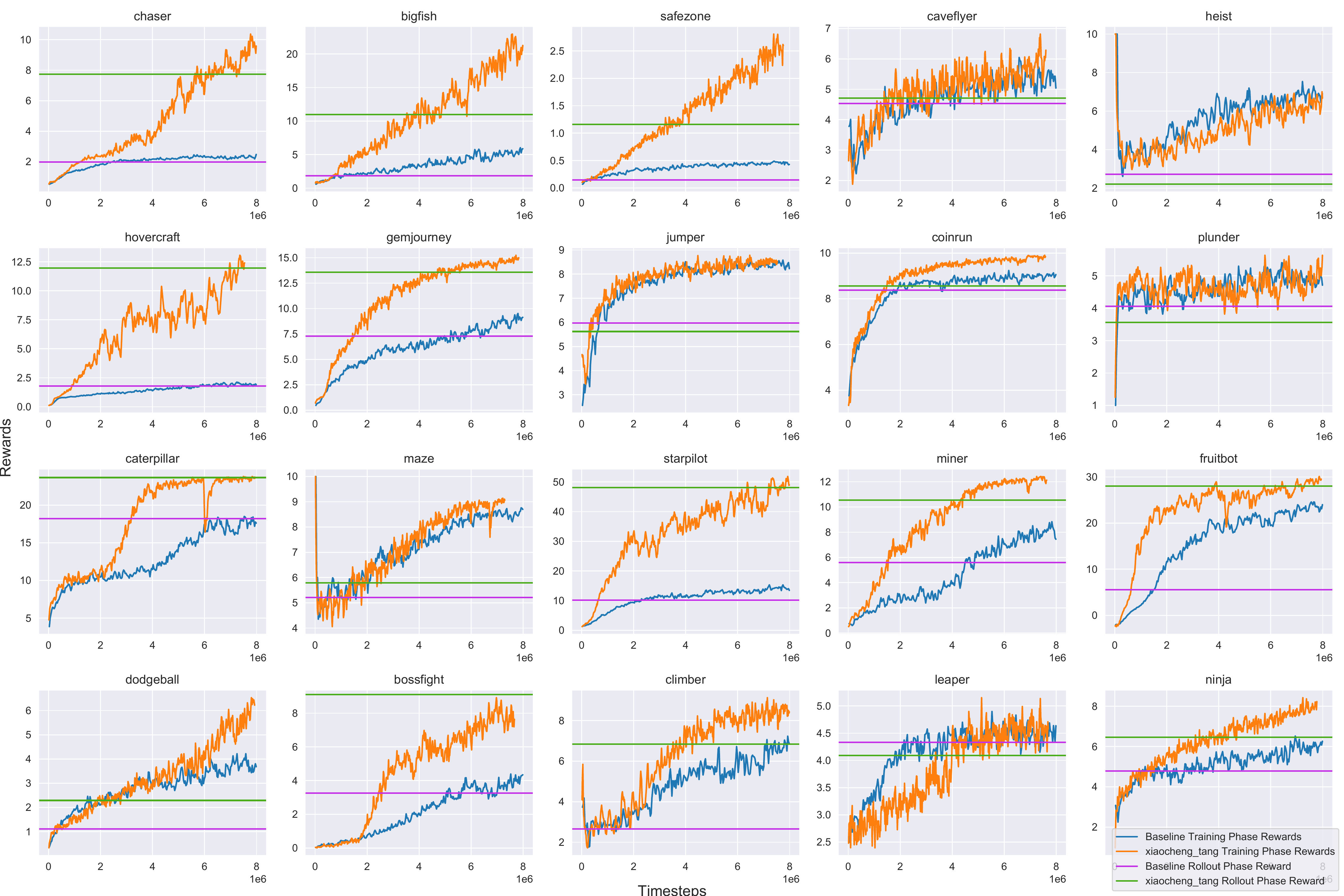}}
\end{figure}

\begin{figure}[htbp]

    \label{fig:zeroGeneralization}%
  {\caption{The mean normalized rewards during training phase across rollouts per environment for Team Zero in generalization track. Note that training score are for the 200 levels, not for the entire distribution of the environment.}}%
  {\includegraphics[width=\textwidth]{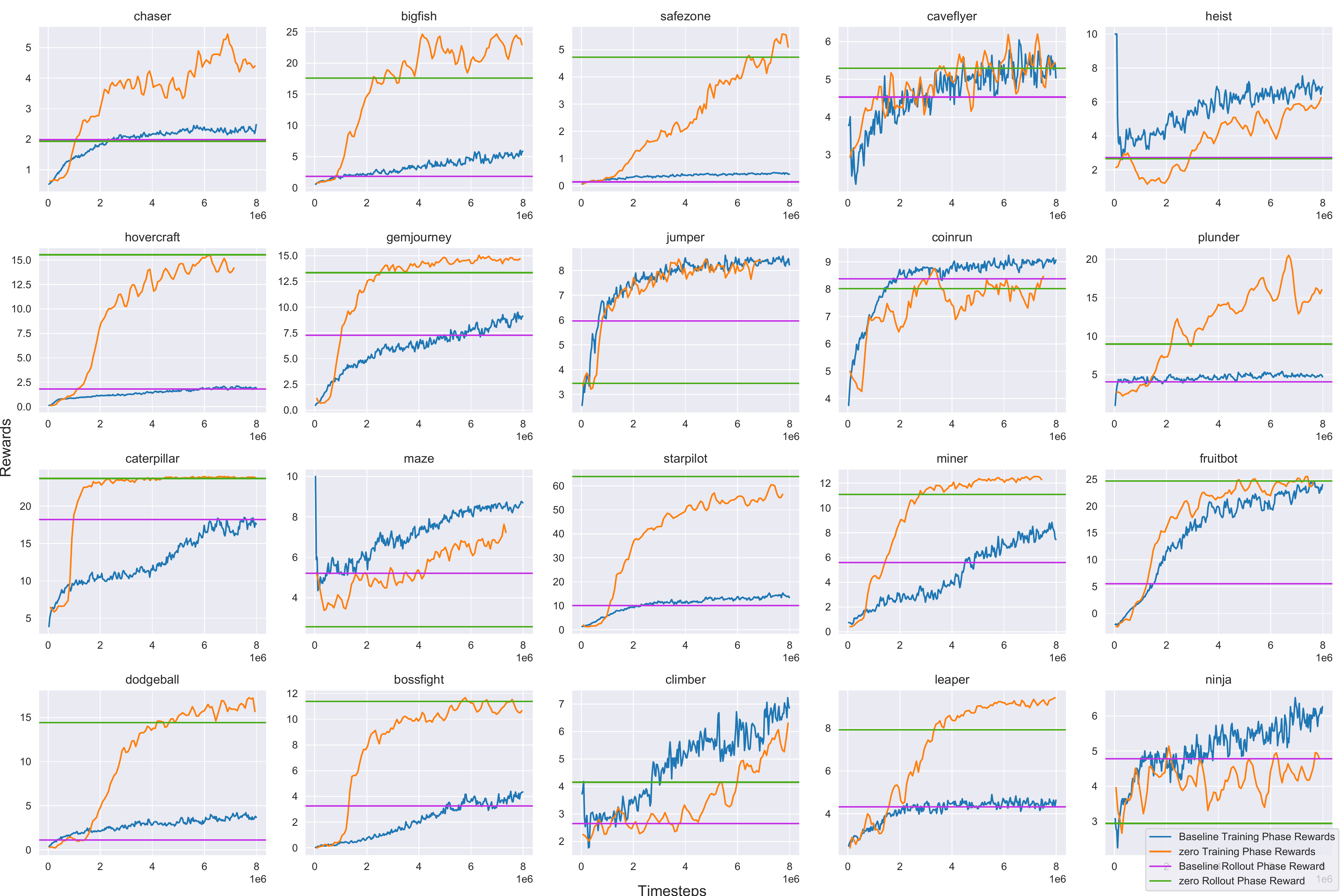}}
\end{figure}

\begin{figure}[htbp]

    \label{fig:paseulGeneralization}%
  {\caption{The mean normalized rewards during training phase across rollouts per environment for Team Paseul in generalization track. Note that training score are for the 200 levels, not for the entire distribution of the environment.}}%
  {\includegraphics[width=\textwidth]{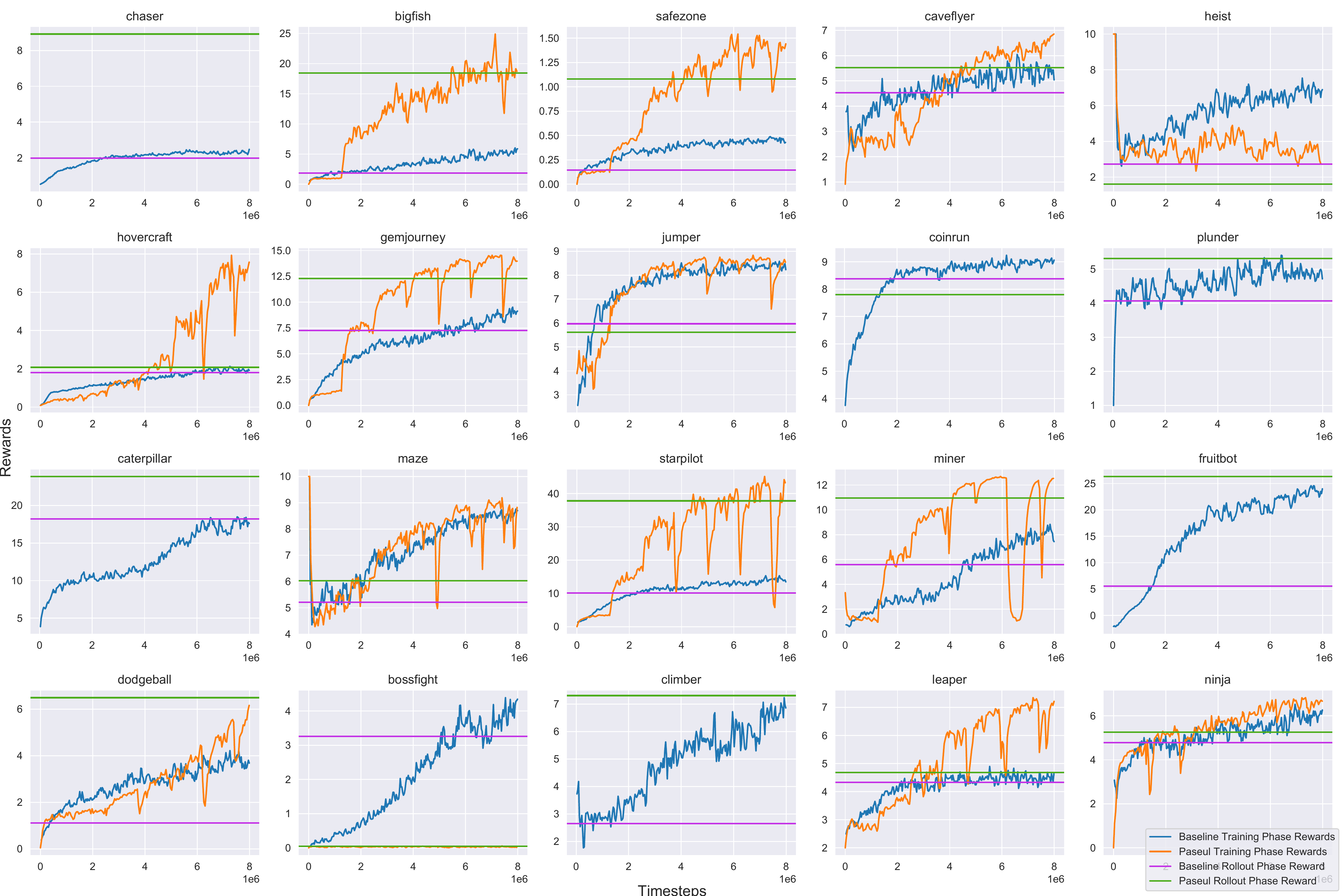}}
\end{figure}

\subsection{Sample-Efficiency Track Graphs}

\begin{figure}[htbp]

    \label{fig:GammaSampleEfficiency}%
  {\caption{The mean normalized rewards across rollouts per environment for Team Gamma in sample-efficiency track.}}%
  {\includegraphics[width=0.99\linewidth]{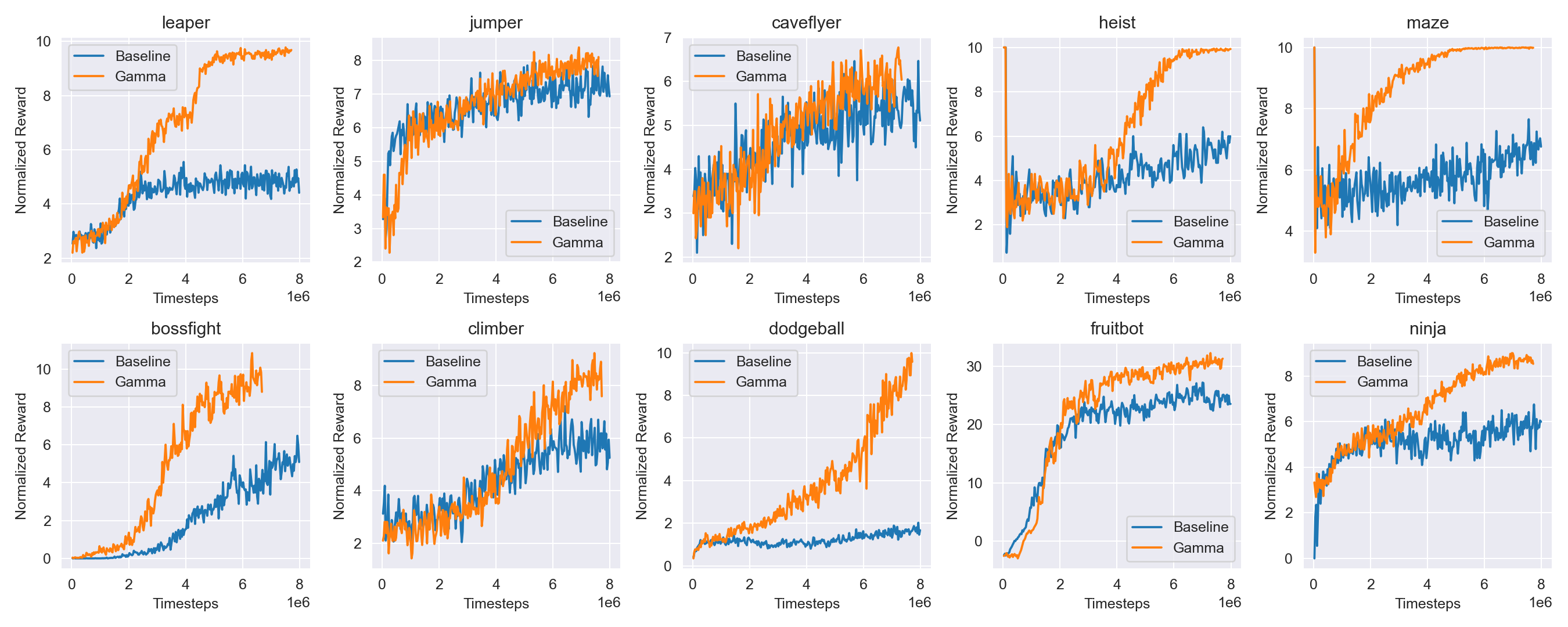}}
\end{figure}

\begin{figure}[htbp]

    \label{fig:wulfebwSampleEfficiency}%
  {\caption{The mean normalized rewards across rollouts per environment for Team TRI in sample-efficiency track.}}%
  {\includegraphics[width=0.99\linewidth]{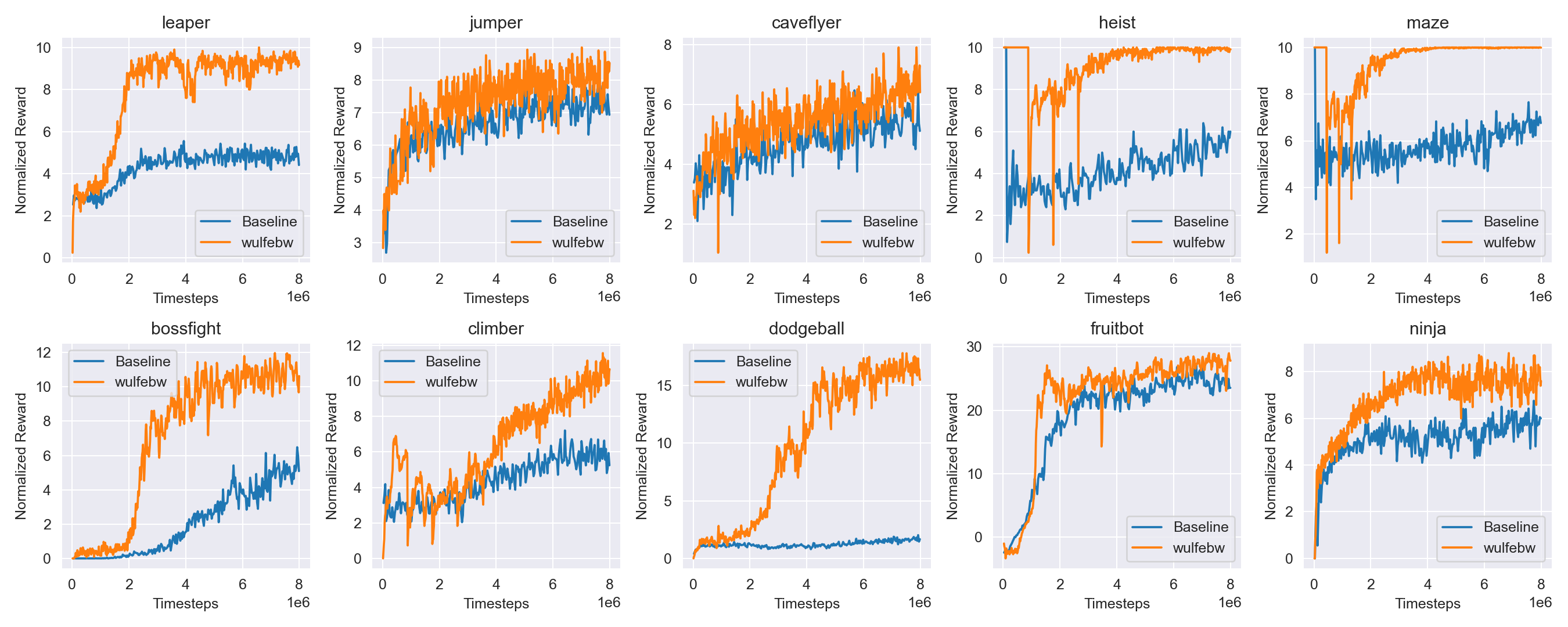}}
\end{figure}

\begin{figure}[htbp]

    \label{fig:AlphaSampleEfficiency}%
  {\caption{The mean normalized rewards across rollouts per environment for Team Alpha in sample-efficiency track.}}%
  {\includegraphics[width=0.99\linewidth]{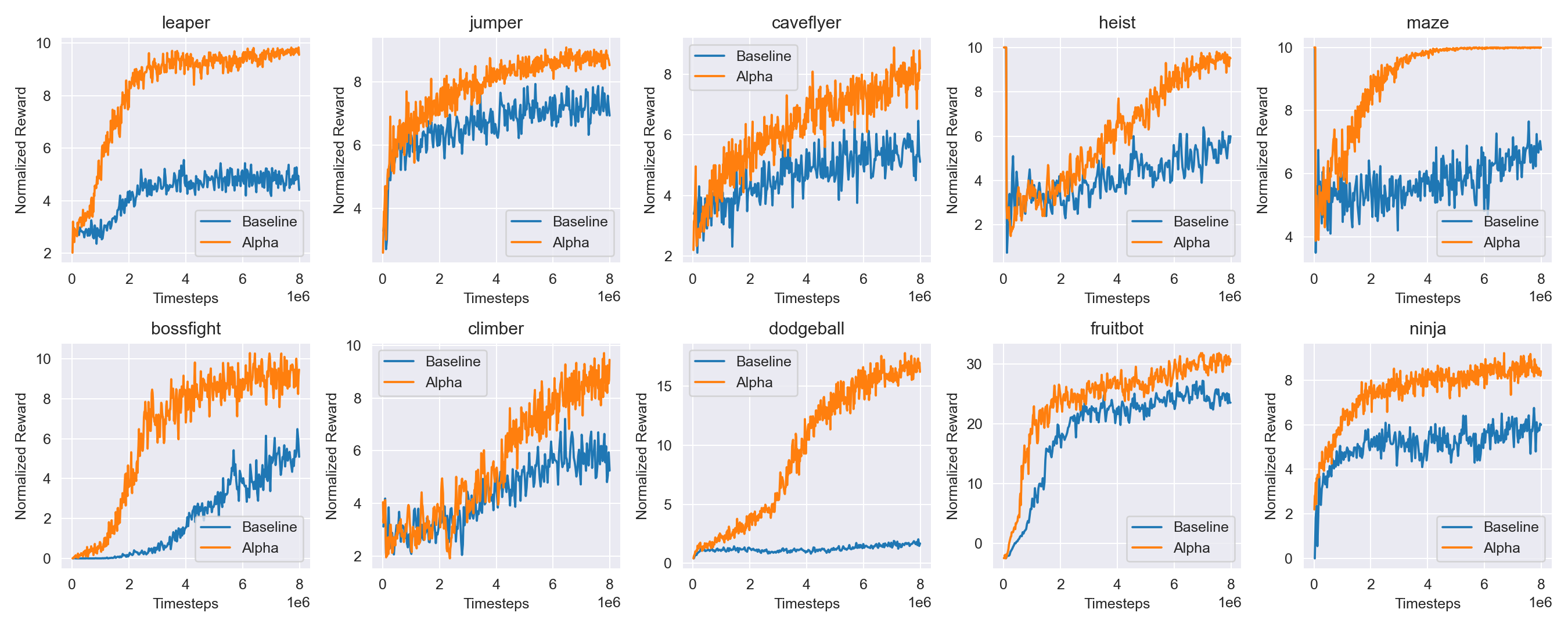}}
\end{figure}

\begin{figure}[htbp]

    \label{fig:MSRLSampleEfficiency}%
  {\caption{The mean normalized rewards across rollouts per environment for Team MSRL in sample-efficiency track.}}%
  {\includegraphics[width=0.99\linewidth]{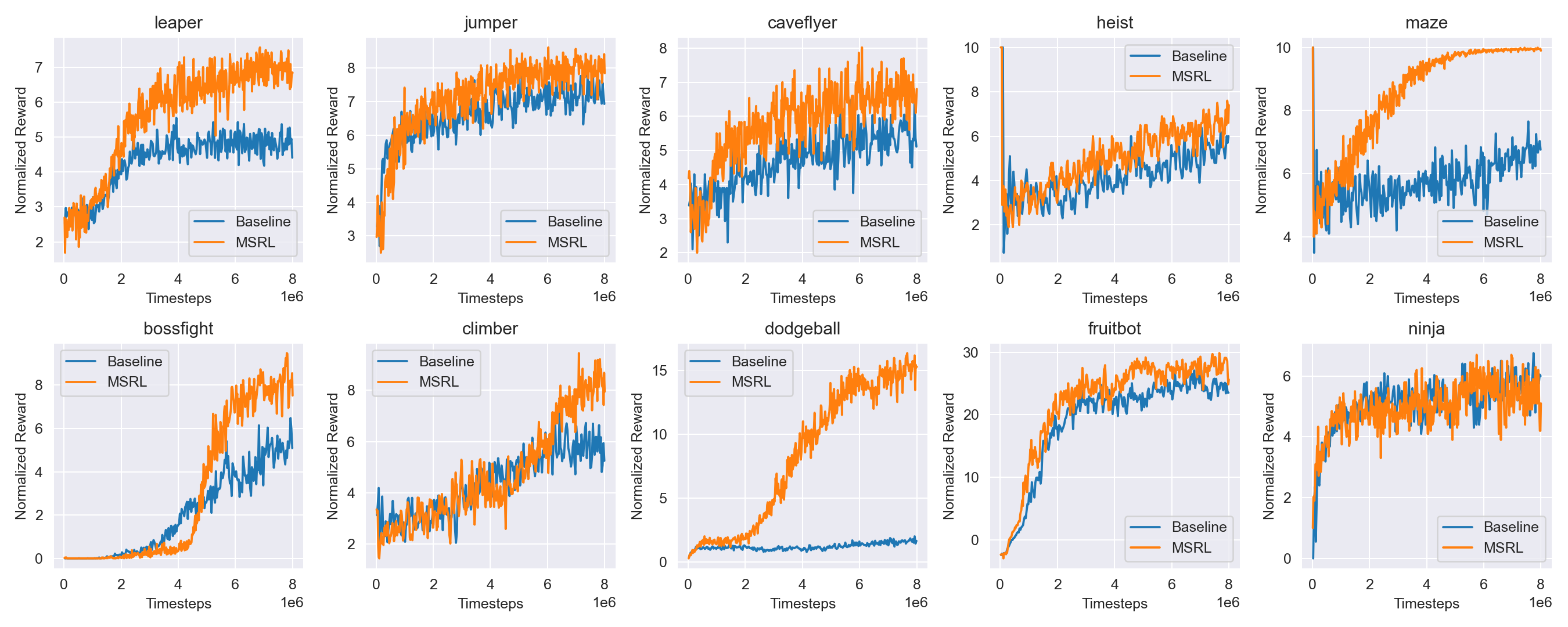}}
\end{figure}

\begin{figure}[htbp]

    \label{fig:three_thirdsSampleEfficiency}%
  {\caption{The mean normalized rewards across rollouts per environment for Team ThreeThirds in sample-efficiency track.}}%
  {\includegraphics[width=0.99\linewidth]{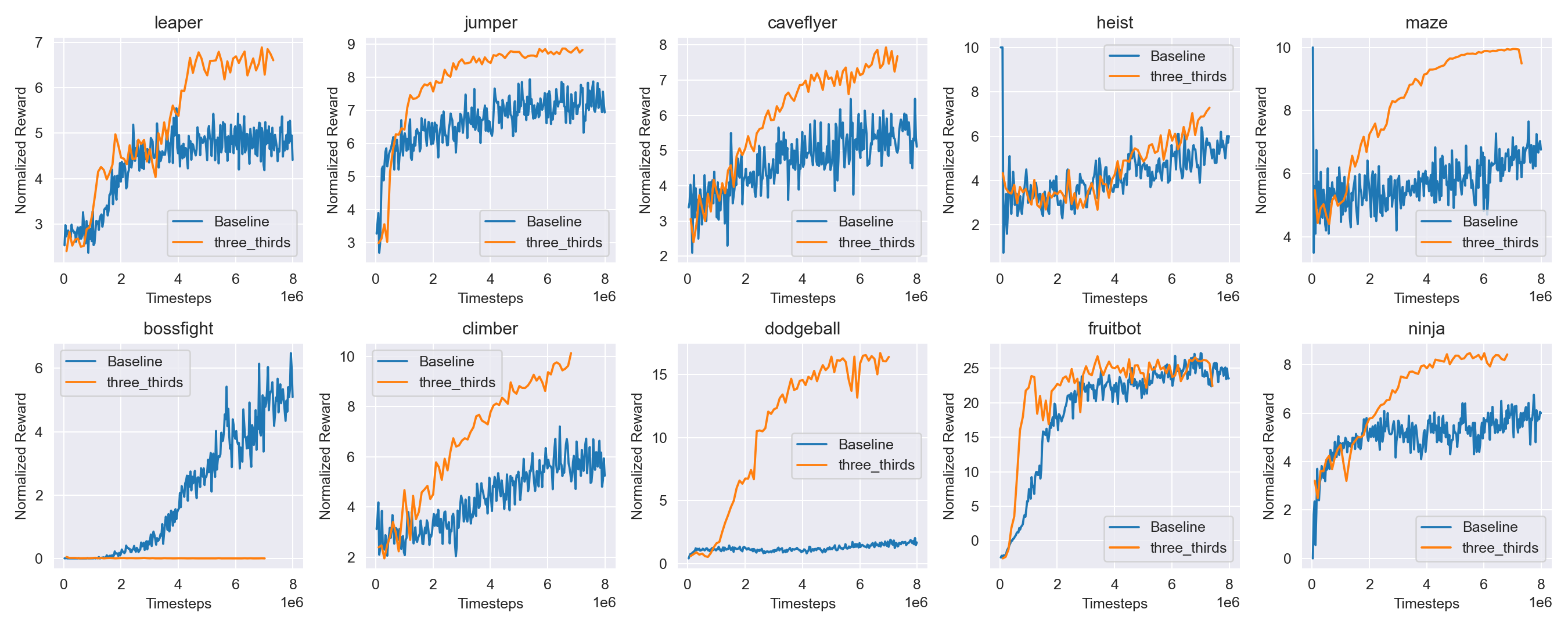}}
\end{figure}

\begin{figure}[htbp]

    \label{fig:JoaoSchapkeSampleEfficiency}%
  {\caption{The mean normalized rewards across rollouts per environment for Individual Joao Schapke in sample-efficiency track.}}%
  {\includegraphics[width=0.99\linewidth]{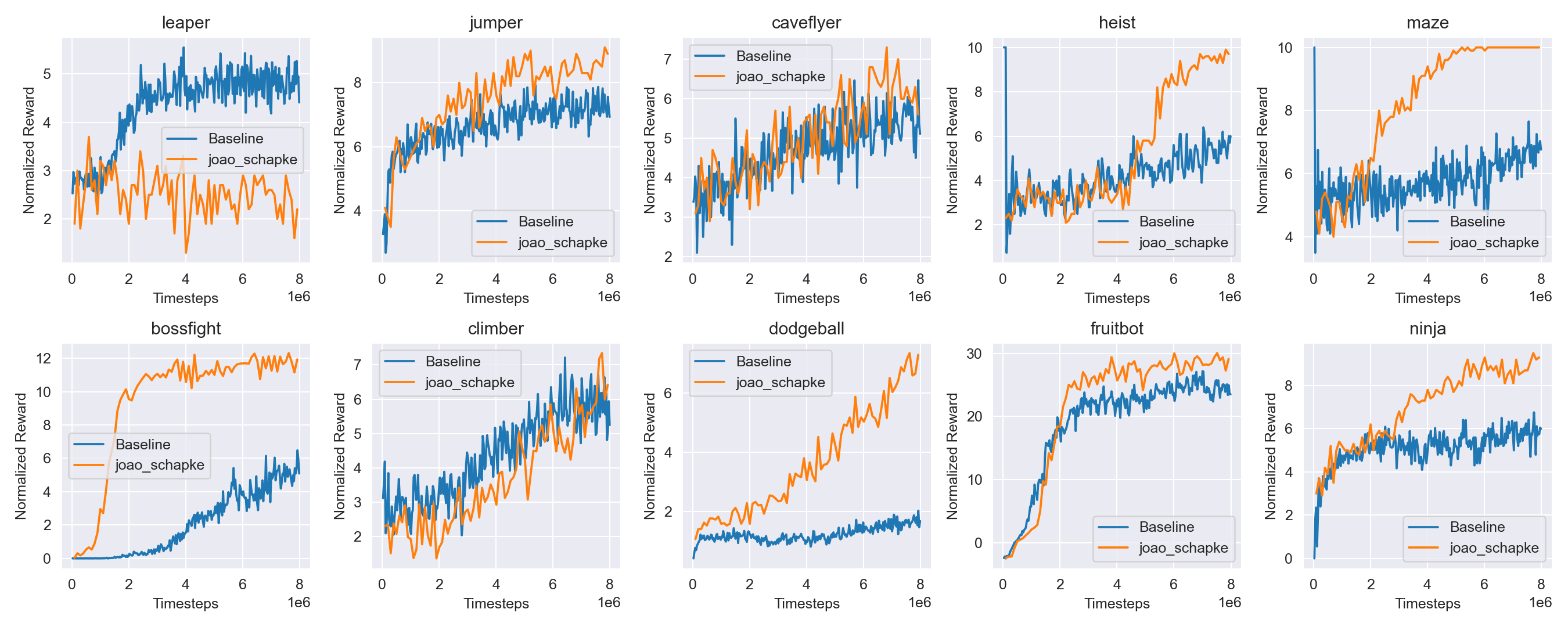}}
\end{figure}

\begin{figure}[htbp]

    \label{fig:ttomSampleEfficiency}%
  {\caption{The mean normalized rewards across rollouts per environment for Individual ttom in sample-efficiency track.}}%
  {\includegraphics[width=0.99\linewidth]{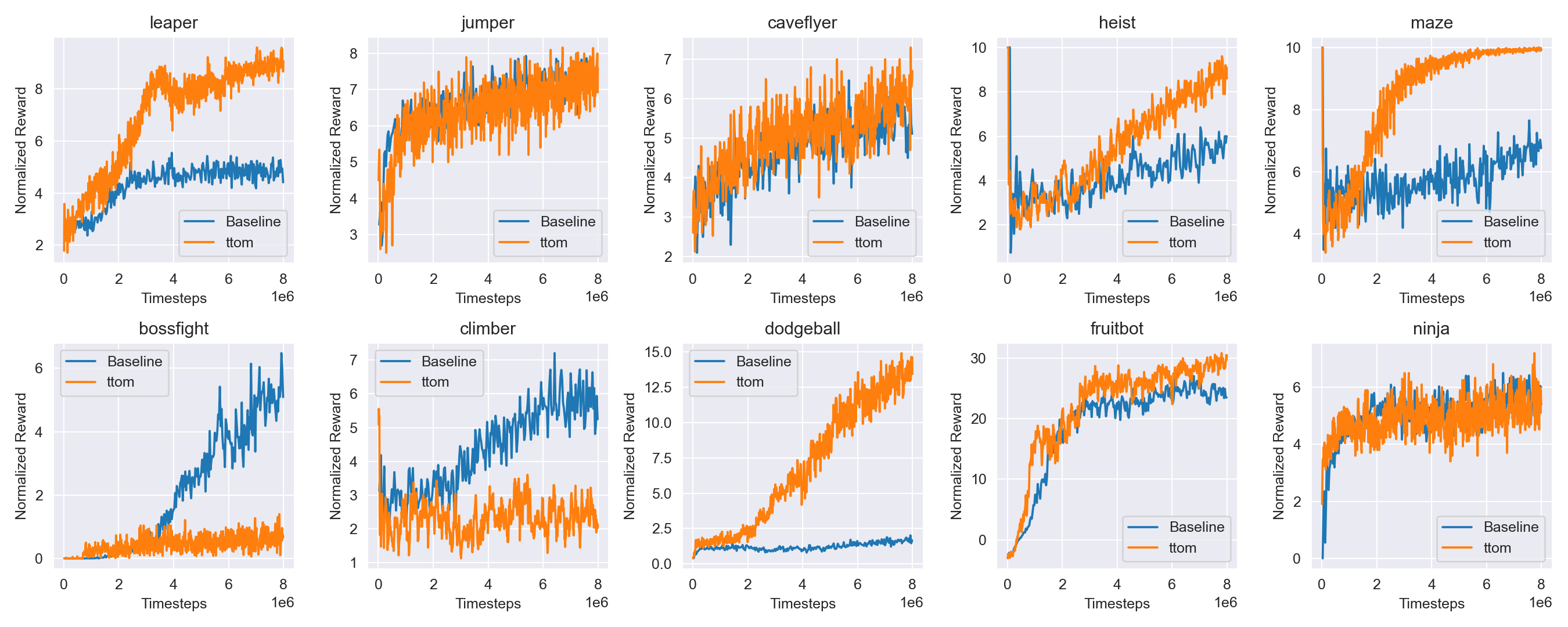}}
\end{figure}

\begin{figure}[htbp]

    \label{fig:XiaochengSampleEfficiency}%
  {\caption{The mean normalized rewards across rollouts per environment for Individual Xiaocheng Tang in sample-efficiency track.}}%
  {\includegraphics[width=0.99\linewidth]{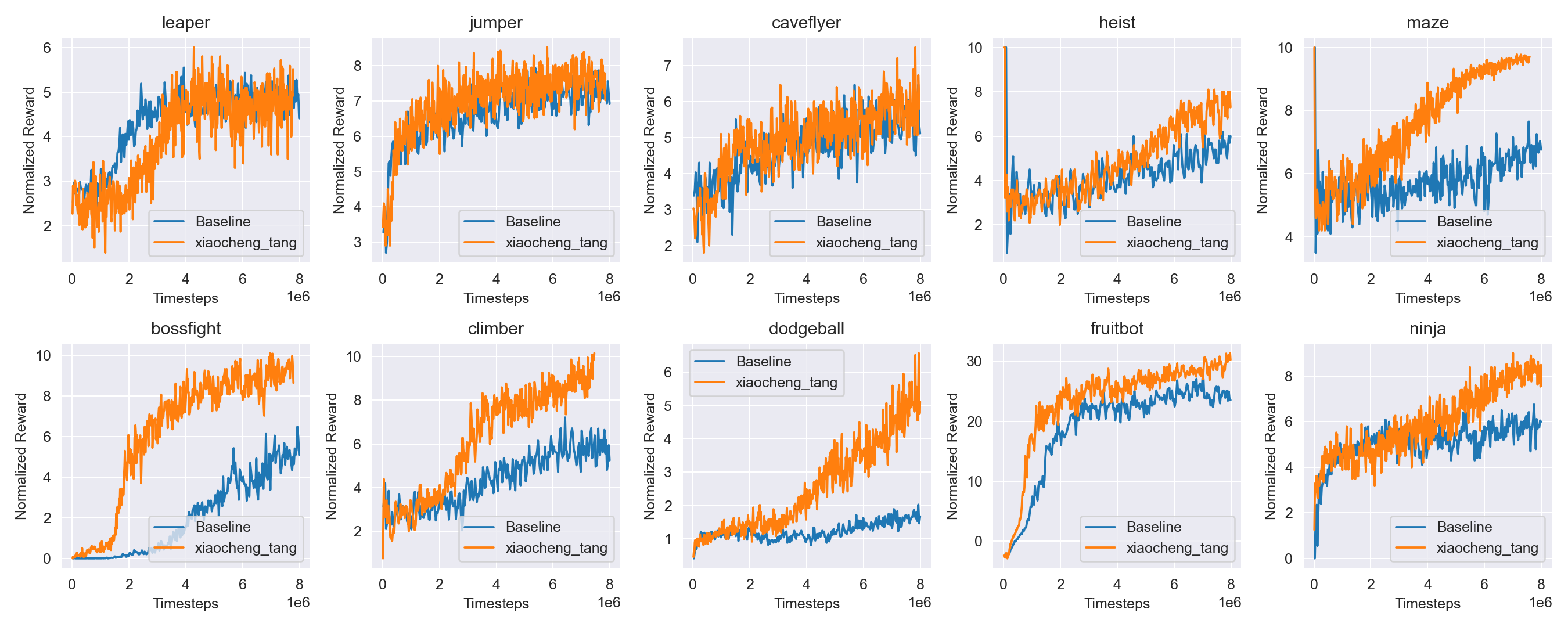}}
\end{figure}

\begin{figure}[htbp]

    \label{fig:zeroSampleEfficiency}%
  {\caption{The mean normalized rewards across rollouts per environment for Team Zero in sample-efficiency track.}}%
  {\includegraphics[width=0.99\linewidth]{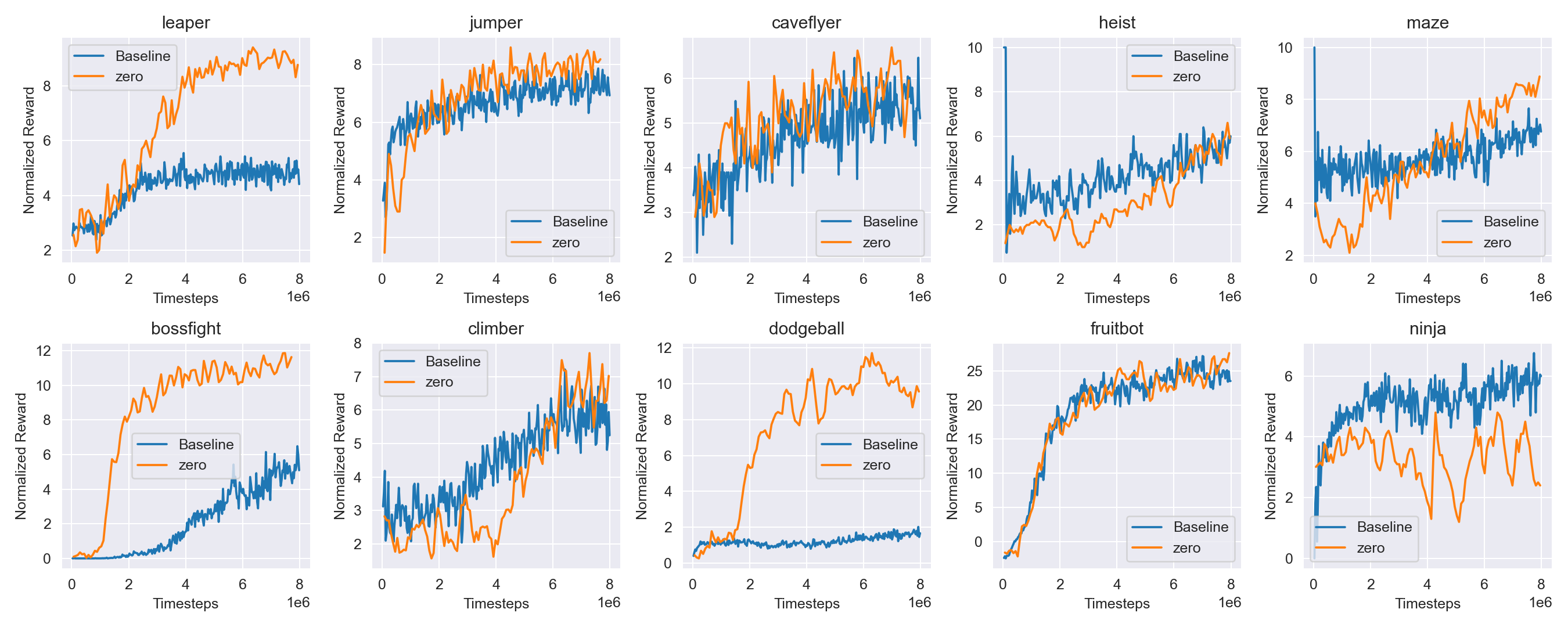}}
\end{figure}

\begin{figure}[htbp]

    \label{fig:paseulSampleEfficiency}%
  {\caption{The mean normalized rewards across rollouts per environment for Team Paseul in sample-efficiency track.}}%
  {\includegraphics[width=0.99\linewidth]{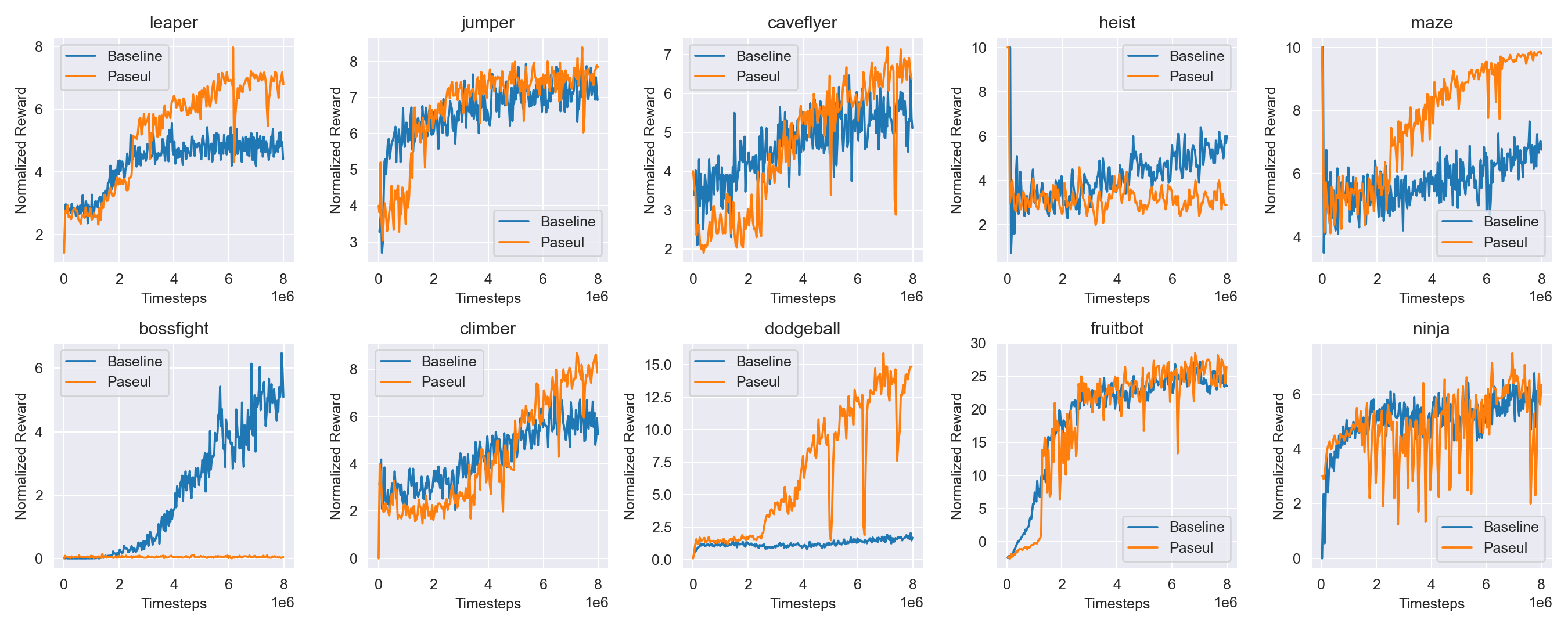}}
\end{figure}

\subsection{Hyper-parameters}

\subsubsection{Team: Gamma}

\begin{tabular}{|r|l|}
        \hline
        Number of Parallel Envs & 
            112 \\ \hline
        Truncated Rollout Length &
            256 \\\hline
        Updates per epoch & 
            8 \\\hline
        PPO Minibatch size & 
            3584 \\\hline
        Frame Stack &
            2 \\\hline
        Aux Phase Frequency & 
            18 \\\hline
        Replay buffer size & 
            500k \\\hline
        $\gamma$ & 
            $ 0.996 $ \\\hline
        $\lambda$ & 
            $ 0.95 $ \\\hline
        Reward Normalization &
            Yes \\\hline
        Replay buffer sampling &
            Uniform Random
            \footnote{ The original PPG implementation samples truncated rollouts instead of single observations} \\
        \hline
    \end{tabular}
\quad
\begin{tabular}{|r|l|}
        \hline
        Impala CNN Depths &
            [32, 64, 64] \\\hline
        Last Dense Layer &
            512 \\\hline
        Preprocessing & 
            Divide by 255 \\\hline
        Optimizer & 
            Adam \\\hline
        Gradient Clipping & 
            None \\\hline
        Auxiliary Epochs &
            7 \\\hline
        Aux Minibatch size &
            2048 \\\hline
        Optimizer Epsilon & 
            $ 1 \times 10^{-8} $ \\\hline
        Learning rate & 
            $ 5 \times 10^{-4} $ \\\hline
        Learning rate Schedule &
            Linear \\\hline
        Final learning rate & 
            $ 5 \times 10^{-5} $ \\
        \hline
    \end{tabular}

\subsubsection{Team: TRI}

\begin{tabular}{|r|l|}
\hline
\textbf{Hyperparameter}                    & \textbf{Value} \\ \hline
Impala Layer Sizes                         & 32, 48, 64     \\ \hline
Rollout Fragment Length                    & 16             \\ \hline
Number of Workers                          & 7              \\ \hline
Number of Envs Per Worker                  & 125            \\ \hline
Minibatch Size                             & 1750           \\ \hline
PPG Auxiliary Phase Frequency              & 32             \\ \hline
PPG Auxiliary Phase Number of Epochs       & 2              \\ \hline
Value Loss Coefficient                     & 0.25           \\ \hline
Framestack                                 & 2              \\ \hline
Dropout Probability (Auxiliary Phase Only) & 0.1            \\ \hline
No-op Penalty                              & -0.1           \\ \hline
\end{tabular}

\subsubsection{Team: MSRL \label{sec:msrl}}

\begin{tabular}{|r|l|}
\hline
\textbf{PPO hyperparameter}                    & \textbf{Value} \\ \hline
Impala layer sizes                         & 32, 64, 64, 128, 128    \\ \hline
Rollout fragment length                    & 256             \\ \hline
Number of workers                          & 2              \\ \hline
Number of environments per worker                  & 64             \\ \hline
Number of CPUs per worker                  & 5             \\ \hline
Number of GPUs per worker                  & 0.1             \\ \hline
Number of training GPUs                 & 0.3             \\ \hline
Discount factor $\gamma$                                & 0.995          \\ \hline
SGD minibatch size                             & 2048           \\ \hline
Batch size                             & 2048           \\ \hline
Number of SGD iterations                & 3            \\ \hline
SGD learning\ rate                          & 0.0006        \\ \hline
Framestacking                           & off           \\ \hline
Truncate episodes              & true             \\ \hline
Value function clip parameter           &1.0            \\ \hline
Value function loss coefficient           &0.5         \\ \hline
Value function share layers             &true           \\ \hline
KL-div. coefficient                    &0               \\ \hline
KL-div. target                    &0.1               \\ \hline
Entropy regularization coefficient      &0.005          \\ \hline
PPO clip parameter                          &0.1           \\ \hline
Gradient norm clip value                           &1          \\ \hline    
GAE  $\lambda$                           &0.8       \\ \hline
L2 regularization coefficient           &0.00001  \\ \hline
\end{tabular}

\subsubsection{Team: ThreeThirds}
   \begin{tabular}{|r|l|}
        \hline
        Environment steps & 
            8M \\\hline
        Generalization training levels &
            200 \\\hline
        Steps trained/sampled & 
            7x \\\hline
        Rollout length (steps) & 
            32 \\\hline
        Batch size (steps) & 
            512 \\\hline
        $\gamma$ & 
            $ 0.995 $ \\\hline
        $\lambda$ & 
            $ 1 $ \\\hline
        Adam learning rate, momentum & 
            $ 1.5 \times 10^{-4}, 0 $ \\\hline
        Target network update freq. (batches) & 
            500 \\\hline
        Replay buffer size & 
            400k \\\hline
        Replay prioritization $\epsilon$ & 
            0.25 \\\hline
        Reward prediction loss coeff. & 
            0.1 \\\hline
        Entropy target: initial, final & 
            2.3, 1.0 \\\hline
        Temperature learning rate & 
            $ 2.5 \times 10^{-4} $ \\\hline
        Random exploration $\varepsilon$ & 
            0.01 \\
        \hline
    \end{tabular}

\subsubsection{Individual: Joao Schapke}
\begin{tabular}{|r|l|}
\hline
\textbf{Hyperparameter}  & \textbf{Value}              \\ \hline
Learning rate & 0.007 \\\hline
entropy\_coeff & 0.0 \\\hline
vf\_coeff & 0.5 \\\hline
gamma & 0.995 \\\hline
gae\_lambda & 0.85 \\\hline
grad\_clip & 1 \\\hline
rollout\_fragment\_length & 32 \\\hline
train\_batch\_size & 2048 \\\hline
\end{tabular}
\quad
\begin{tabular}{|r|l|}
\hline
\textbf{Hyperparameter (P3O)}  & \textbf{Value}              \\ \hline
buffer\_size & 20000 \\\hline
learning\_starts & 5000 \\\hline
prioritized\_replay\_alpha & 2 \\\hline
prioritized\_replay\_beta & 1 \\\hline
prioritized\_replay\_eps & 0.0000001 \\\hline
\end{tabular}

\subsubsection{Individual: ttom}

\begin{tabular}{|r|l|}
\hline
\textbf{Hyperparameter}  & \textbf{Value}              \\ \hline
Truncated Rollout Length & 200                         \\\hline
Updates per epoch        & 3                           \\\hline
PPO Minibatch size       & 1024                        \\\hline
Frame Stack              & 2 (modified)                \\\hline
$\gamma$                 & 0.99                        \\\hline
$\lambda$                & 0.9                         \\\hline
Reward Normalization     & No                          \\\hline
Impala CNN Depths        & {[}32, 64, 64{]}            \\\hline
Last Dense Layer         & 256                         \\\hline
Preprocessing            & Subtract 128, divide by 255 \\\hline
Optimizer                & Adam                        \\\hline
Gradient Clipping        & None                        \\\hline
Learning rate            & 1E-4-\textgreater{}3E-5     \\\hline
Learning rate schedule   & Cosine \\\hline
\end{tabular}

\end{document}